\documentclass[11pt]{article}

\usepackage[final]{acl}

\usepackage{times}
\usepackage{latexsym}

\usepackage[T1]{fontenc}

\usepackage[utf8]{inputenc}

\usepackage{microtype}

\usepackage{inconsolata}

\usepackage{graphicx}

\title{Can LLMs Model Incorrect Student Reasoning? \\
A Case Study on Distractor Generation for Multiple-Choice Questions}

\author{
  \textbf{Yanick Zengaffinen\textsuperscript{\textnormal{1}}}
  \textbf{Andreas Opedal\textsuperscript{\textnormal{1}}}
  \textbf{Donya Rooein\textsuperscript{\textnormal{2}}}
  \textbf{Kv Aditya Srivatsa\textsuperscript{\textnormal{1}}}\\
  \textbf{Shashank Sonkar\textsuperscript{\textnormal{3}}}
  \textbf{Mrinmaya Sachan\textsuperscript{\textnormal{1}}}
\\
\\
  \textsuperscript{1}ETH Zürich\;
  \textsuperscript{2}Bocconi University\;
  \textsuperscript{3}University of Central Florida
\\
\texttt{\href{mailto:yanick.zengaffinen@inf.ethz.ch}{yanick.zengaffinen@inf.ethz.ch}}
}

\usepackage{multirow}
\usepackage{amsmath} 
\usepackage{hyperref}       
\usepackage{cleveref}
\usepackage{url}            
\usepackage{booktabs}       
\usepackage{graphicx}
\usepackage{tabularx}
\usepackage{enumitem}
\usepackage{bm}
\usepackage{xspace}
\usepackage{amssymb}
\usepackage{placeins}

\usepackage{algorithm}
\usepackage{algpseudocode}

\usepackage[most]{tcolorbox}

\crefname{figure}{Fig.}{Figs.}
\crefname{table}{Table}{Tables}
\crefname{appendix}{App.}{Apps.}
\crefname{section}{\S}{\S\S}
\crefformat{section}{\S#2#1#3}
\crefname{equation}{Eq.}{Eqs.}
\crefname{algorithm}{Alg.}{Algs.}
\crefname{algocf}{Alg.}{Algs.}
\crefname{defin}{Def.}{Defs.}
\crefname{theorem}{Thm.}{Thms.}
\crefname{lemma}{Lemma}{Lemmas}

\newcommand{\defn}[1]{\textbf{#1}}

\newlist{dialogue}{description}{1}
\setlist[dialogue]{
    labelwidth=1.5cm,
    labelindent=0cm,
    leftmargin=1.8cm,
    labelsep=0.3cm,
    align=left,
    noitemsep,
    topsep=0pt
}
\newcommand{\promptvar}[1]{\textit{\textcolor{gray}{#1}}}

\definecolor{colInter}{HTML}{1F77B4}
\definecolor{colCorr}{HTML}{2CA02C}
\definecolor{colErrDesc}{HTML}{FF7F0E}
\definecolor{colInst}{HTML}{D62728}
\definecolor{colPlaus}{HTML}{9467BD}
\definecolor{colCur}{HTML}{17BECF}

\newcommand{\ds}{\textsc{DeepSeek-V3.2}\xspace}
\newcommand{\dsabbr}{\textsc{DS-V3.2}\xspace}
\newcommand{\glm}{\textsc{GLM-4.7}\xspace}
\newcommand{\gptoss}{\textsc{GPT-oss-20b}\xspace}
\newcommand{\gptfourone}{\textsc{GPT-4.1}\xspace}
\newcommand{\gptfouronemini}{\textsc{GPT-4.1-mini}\xspace}

\newcommand{\gptfive}{\textsc{GPT-5}\xspace}
\newcommand{\gptfivetwo}{\textsc{GPT-5.2}\xspace}
\newcommand{\gptosslarge}{\textsc{GPT-oss-120b}\xspace}
\newcommand{\gemma}{\textsc{Gemma-3}\xspace}
\newcommand{\geminis}{\textsc{Gemini-2.5 (flash/pro)}\xspace}
\newcommand{\glmflash}{\textsc{GLM-4.7 flash}\xspace}

\makeatletter
\newcommand{\oset}[3][0.23ex]{%
  \mathrel{\mathop{#3}\limits^{
    \vbox to#1{\kern-2\ex@
    \hbox{$\scriptstyle#2$}\vss}}}}
\makeatother
\newcommand{\defeq}[0]{\oset[0.4ex]{\text{\tiny def}}{=}}

\begin{document}
\maketitle
\begin{abstract}
Modeling student misconceptions in a realistic manner is critical for AI in education.
In this work, we examine how large language models (LLMs) reason about misconceptions when generating distractor answers for multiple-choice questions (MCQs), a task that requires producing answers that are incorrect, yet plausible.
We introduce a taxonomy over reasoning strategies for distractor generation that is grounded in learning-science literature and empirical observation, which we apply to LLM-generated reasoning traces across math and science MCQs.
On the math dataset, we find that models follow a misconception-based process with potentially high diagnostic value: they recover the correct solution, articulate student errors, simulate them, and select plausible candidates.
On the science dataset, on the other hand, they tend to follow a less robust approach based on semantic similarity to the correct answer.
We find the most frequent failure modes to be that the model is unable to generate a correct solution or that it discards plausible distractor candidates when performing selection.
Providing the correct solution in the prompt yields a relative improvement of 6.4\% in alignment with human-authored distractors, highlighting the critical role of anchoring distractor generation to the correct solution.
Together, our findings offer an interpretable view of how LLMs model incorrect student reasoning.\footnote{All code is available on GitHub at \url{https://github.com/eth-lre/llm-student-modeling-strategies}.}
\end{abstract}

\section{Introduction}
Student modeling is a foundational problem in AI for education~\citep{Matsuda07,Kaser2023SimulatedLI}, underpinning a range of applications such as test design~\citep{benedetto-etal-2024-using,Ezzaki25}, misconception diagnosis, teacher training~\citep{judge2013use,leegenerative} and development of targeted instructional interventions \citep{Jin25,rooein2026pats}. A central component of student modeling is the ability to represent not only what students know, but also the systematic errors and misconceptions that characterize incorrect but plausible reasoning. Despite rapid progress in large language model (LLM) capabilities
and a growing focus on using LLMs for student simulation \citep{macina-etal-2023-mathdial,opedal2024cognitive,liu2025llmsmakemistakeslike,Wu2025EmbracingISA}, existing work has largely studied LLM alignment with student behavior by evaluating whether they arrive at similar final answers,
leaving open the complementary question of \emph{how} LLMs actually simulate student mistakes.

\begin{figure*}[t]
\centering

\includegraphics[width=0.8\linewidth]{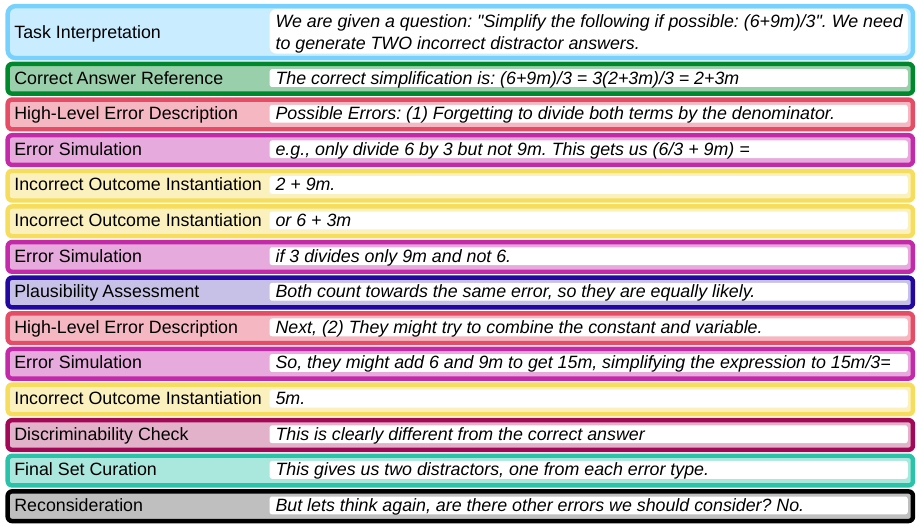}

\caption{
An LLM reasoning trace for distractor generation, annotated according to our taxonomy (\cref{tab:taxonomy-definition}).
Color-coded labels show the model interpreting the task, anchoring to the correct solution, generating candidate distractors based on errors, evaluating their plausibility, and curating the final set. This mirrors strategies identified in the learning-science literature (\cref{sec:ls-foundations}).
The \texttt{LINK} strategy, which captures conceptual similarity to the correct answer, does not appear in this trace, as it is rarely used on the math dataset (\cref{sec:experiments-process-analysis}). In subsequent traces, we request three distractors rather than the two shown here.
}

\label{fig:main-annotation-example}
\end{figure*}

To answer this question, we study distractor generation, the task of generating incorrect but plausible answer alternatives, i.e., \defn{distractors}, for multiple-choice questions (MCQs). Distractor generation is a natural application of student modeling: if a model can faithfully simulate plausible student errors, distractors can be obtained by sampling from simulated incorrect solutions and selecting probable outcomes. To produce good distractors, a model is required not only to identify the correct solution, but also to simulate \emph{incorrect} student reasoning and to verify the plausibility of the resulting answers~\citep{gierl2017distractorreview,haladyna2013developing}.
Distractor generation is therefore a strictly more difficult task than the one of \emph{solving} problems.
While prior work has investigated using LLMs for distractor generation~\citep{feng-etal-2024-exploring,alhazmi2024distractorgenerationmultiplechoicetasks,doughty2024compstudy,sauberli-clematide-2024-automatic,parikh-etal-2025-lookalike}, the focus has largely been on evaluating final answers. These alone do not reveal whether distractors arise from pedagogically meaningful reasoning, or something more akin to surface-level heuristics. In educational settings, it is particularly important to understand the underlying process, both for aligning LLM strategies with expert practices and for building teacher trust in AI-based tools \citep{nazaretsky2022trust,Feldman-Maggor2025}.\looseness=-1

We propose a framework for analyzing the reasoning traces generated by LLMs during distractor generation.\footnote{Note that we do not study reconstruction of any real student's reasoning; accordingly, we assess the reasoning traces against established pedagogical theory (misconception-based item design) rather than cognitive data from real students.}
Each such reasoning \defn{trace} corresponds to a single LLM generating the distractors for one MCQ.
We construct a learning-science-aligned taxonomy of ten distractor-generation strategies, defined in \cref{tab:taxonomy-definition} and grounded in principles of misconception-based assessment design. We complement the taxonomy with empirical observation through iterative open coding of LLM reasoning traces \citep{corbin_strauss_2008_basics}, a qualitative process for identifying and labeling recurring concepts in data.
The taxonomy provides a theory-grounded lens for systematically characterizing and diagnosing \emph{how} LLMs generate distractors, rather than only evaluating the final distractor set.
We apply this taxonomy to 885 distractor-generation traces from \ds~\citep{deepseekai2025deepseekv32pushingfrontieropen} and \glm~\citep{5team2025glm45agenticreasoningcoding} across two MCQ datasets spanning mathematics and science.
\cref{fig:main-annotation-example} illustrates the annotation of a single reasoning trace according to the strategies in our taxonomy.

Most notably, we find that on the misconception-annotated math dataset, LLM reasoning traces show close alignment with best practices for distractor design \citep{haladyna2013developing,gierl2017distractorreview}: models recover the correct solution, articulate plausible student misconceptions, instantiate them through erroneous reasoning, and select plausible incorrect answers from a broader candidate set. The recovered solution serves as an anchor from which the model diverges at specific points to generate misconception-aligned distractors. On the science dataset, however, models rely more heavily on superficial semantic similarity to the correct answer: they recover the correct answer, establish a conceptual link to it, generate candidates along that link, and verify that these are indeed incorrect before selecting a final set.
Across all settings, we observe a consistent high-level structure for problem understanding, candidate distractor construction, evaluation, and selection.

Beyond characterizing LLMs' processes, our taxonomy also enables a diagnosis of where and why distractor generation fails. We find that some parts rarely fail: both error simulation, i.e., simulating a student's flawed reasoning step by step, and the implicit organization of multi-step reasoning, i.e., the order in which strategies are chained. Note that the latter emerges without instruction: the model self-structures it. Instead, failures arise more often in recovering the correct solution and in the downstream selection of the final distractor set. Providing the correct solution in the prompt yields significant relative improvements in the fraction of human-authored distractors the model recovers (proportional match) of 6.4\% on the math dataset and 30.7\% on the science dataset,\footnote{\label{fn:sciq-corr-improve-caveat}The larger relative gain on the science dataset should be interpreted with caution given the absolute improvement (0.14 $\rightarrow$ 0.18) is small relative to the confidence intervals ($\pm$0.02 and 0.03). Both improvements are significant under a paired $t$-test, $p<0.01$.} confirming that unreliable solution recovery is a primary bottleneck.
We also find that models discard plausible candidates during selection, indicating that candidate ranking is another source of performance degradation.

Overall, our results show that LLMs can systematically engage in misconception-based distractor generation while adapting their strategies across domains, with correct-solution recovery serving as a key anchor.
Together, these findings deepen our understanding of when and how modern LLMs can model plausible incorrect student reasoning.

\section{Related Work}
\paragraph{Student modeling with LLMs.}
A growing body of work suggests LLMs as promising tools for simulating student behavior. They can generate student responses that expert annotators judge as reflective of common grade-appropriate misconceptions \citep{macina-etal-2023-mathdial} and condition their behavior on a learner's cognitive level \citep{Wu2025EmbracingISA} or misconceptions \citep{Liu2023NoviceLAA}.
At the same time, recent literature suggests that LLMs may encode common student error patterns \citep{opedal2024cognitive,liu2025llmsmakemistakeslike}, but it remains unclear whether they can effectively leverage this implicit knowledge for student simulation.
While these findings suggest that LLMs can simulate incorrect student reasoning to some extent, prior evaluations focus primarily on final outputs, rather than the reasoning processes that produce them.
As a result, it remains unclear which reasoning strategies LLMs use to generate incorrect answers.\looseness=-1

\paragraph{Multiple-choice questions and distractor generation.}
MCQs are widely used to assess student understanding \citep{scouller1998influence,chingos2012strength}, but their effectiveness depends on the quality of the distractors.
Crafting high-quality distractors (see \cref{sec:ls-foundations}) is costly and time consuming \citep{rudner2010adaptive,quaigrain2017}. Prior work has therefore explored automating this process with LLMs \citep[\emph{inter alia}]{alhazmi2024distractorgenerationmultiplechoicetasks}, particularly for mathematical MCQs \citep{feng-etal-2024-exploring,hang2024mcqgen,lee2024mathmultiplechoicequestion,parikh-etal-2025-lookalike}. Some approaches decompose distractor generation into producing candidate mistakes and applying them to generate plausible answers, reflecting the misconception-based approach formalized in \cref{sec:ls-foundations} \citep{fernandez-etal-2024-divert,ross2025learningmakemistakesmodeling}, while others rely on overgeneration followed by ranking \citep{scarlatos-etal-2024-improving}.
Despite notable performance gains, because this line of work measures only the final output set, the strategies underlying LLM-based distractor generation and their relation to learning-science principles remain insufficiently understood. We therefore complement it with a process-level analysis of the reasoning traces themselves, localizing failures to specific stages of the generation process.

\section{A Taxonomy of Strategies for Distractor Generation}
Our goal is to characterize the strategies LLMs use to generate distractors.
We develop a taxonomy, shown in \cref{tab:taxonomy-definition}, by combining learning-science theory with empirical analysis to capture both expected strategies and observed LLM behavior.
We first review experts' strategies (\cref{sec:ls-foundations}), then describe how we construct our taxonomy (\cref{sec:taxonomy-method-details}), and finally present the annotation pipeline used to apply it to LLM traces (\cref{sec:method-annotation}).

\subsection{Crafting Effective Distractors}
\label{sec:ls-foundations}
Effective distractors must satisfy multiple criteria: they must be plausible, i.e., likely to be selected by students with flawed or incomplete understanding, be unambiguously incorrect, and not contain cues such as grammatical inconsistencies. In addition, a \emph{set} of distractors must be mutually exclusive and avoid structural artifacts that reveal the correct answer~\citep{haladyna2013developing,gierl2017distractorreview,tarrant2009functioning,haladyna2002}.
Learning-science literature distinguishes two main construction pipelines.
\defn{Similarity-based} distractors are generated by perturbing the surface form or semantic properties of the correct answer without modeling student reasoning~\citep{chamberlain2020integrity}. For example, if the correct answer is $\frac{4}{3}$, a distractor could invert the sign to $-\frac{4}{3}$.
\defn{Misconception-based} distractors, on the other hand, are derived from common student errors injected into an otherwise correct solution process~\citep{brown1980repair,vanlehn1990mind,shin2019multiple}.
For example, if students forget to invert the divisor when dividing fractions, a distractor can be generated by applying such an error (e.g., $\frac{1}{3} \div \frac{1}{2} \overset{!}{=} \frac{1}{6}$).
Misconception-based distractors are more costly and complex to construct than similarity-based ones~\citep{gierl2017distractorreview}, but are often preferred due to higher diagnostic value, as they expose specific procedural gaps in students' understanding~\citep{ashlock2010error,delatorre2009mcdina}.\looseness=-1

\begin{table*}[!htbp]
\centering

\vspace{-5pt}
\small
\begin{tabular}{rl|p{0.62\textwidth}}
 \toprule
 & \textbf{Strategy} & \textbf{Definition} \\
 \midrule
 \multicolumn{3}{l}{\textit{Understanding}} \\
 \texttt{INTER}    & Task Interpretation &
   Clarifying the problem statement / instructions to determine the required output. \\
 \texttt{CORR}     & Correct Answer Ref. &
   Retrieving, deducing, or justifying the correct answer to the MCQ. \\
 \midrule
 \multicolumn{3}{l}{\textit{Candidate Construction}} \\
 \texttt{LINK}     & Conceptual Link &
   Establishing the relation distractors should share with the correct answer. \\
 \texttt{ERR\_DESC}& Error Description &
   Abstract description of a learner action or misconception (``a student might\ldots''). \\
 \texttt{ERR\_SIM} & Error Simulation &
   Step-by-step simulation of a learner's flawed reasoning. \\
 \texttt{INST}     & Outcome Instantiation &
   Any mention of a candidate distractor or incorrect answer. \\
 \midrule
 \multicolumn{3}{l}{\textit{Quality Control and Selection}} \\
 \texttt{PLAUS}    & Plausibility Check &
   Assessing if or why a candidate would be tempting/believable to a student. \\
 \texttt{DISCR}    & Discriminability Check &
   Evaluating whether a candidate is clearly incorrect and unambiguous. \\
 \texttt{CURATE}   & Final Set Curation &
   Selecting, refining, or organizing candidates to improve overall set quality. \\
 \midrule
 \multicolumn{3}{l}{\textit{Reconsideration (Cross-cutting)}} \\
 \texttt{RECON}    & Reconsideration &
   Challenging or revisiting a previous assumption (``but wait\ldots''). \\
 \bottomrule
 \end{tabular}
\caption{A taxonomy for distractor generation grounded in learning-science literature and empirical observations of LLMs' reasoning traces, which we use to annotate reasoning traces as shown in \cref{fig:main-annotation-example}. Appendix~\cref{tab:strategy-examples} illustrates each strategy with trace excerpts.}
\label{tab:taxonomy-definition}
\vspace{-5pt}
\end{table*}

\subsection{Taxonomy Construction}
\label{sec:taxonomy-method-details}
To develop a taxonomy that integrates expectations derived from expert practices (\cref{sec:ls-foundations}) with patterns observed in LLM distractor-generation traces, we first identify expected strategies from the literature, then extract strategies from actual LLM traces, and finally merge the two sets into a unified taxonomy.

\paragraph{Grounding in theory.}
From the approaches described above, we derive the following strategies, assigning each a tag (in parentheses) for reference throughout the paper.
Both similarity-based and misconception-based approaches require access to the correct solution (\texttt{CORR}) and eventually instantiate concrete distractor values (\texttt{INST}).
While similarity-based methods require establishing a semantic or conceptual link to the correct answer (\texttt{LINK}), misconception-based distractor generation instead involves identifying a common error (\texttt{ERR\_DESC}) and simulating this error within an otherwise correct step-by-step solution procedure (\texttt{ERR\_SIM}). Item-level quality constraints correspond to plausibility and discriminability checks (\texttt{PLAUS}, \texttt{DISCR}), while set-level constraints require explicit selection or curation over candidate distractors (\texttt{CURATE}).\looseness=-1

\paragraph{Grounding in observation.}
To ensure our taxonomy represents actual, probable LLM reasoning, we collect and analyze reasoning traces from multiple models and datasets spanning different domains. We use a prompt that avoids biasing models toward any particular strategy; see Appendix~\cref{app:prompts-collect}.
Each trace corresponds to one LLM generating distractors for one MCQ.
To identify the strategies in LLMs' reasoning, we perform manual open coding~\citep{corbin_strauss_2008_basics}, a qualitative method for breaking text into discrete concepts without predefined categories, on a stratified subset of 48 traces across datasets, models, and levels of alignment with human-annotated distractors.
To further broaden coverage, we prompt \ds to perform open coding on groups of 10 traces and report recurring strategies across each group. This is applied to 800 additional traces, without providing any previously extracted strategies.
Experimental details are provided in~\cref{sec:experimental-setup}.

\paragraph{Merging theory and observation.}
Finally, we manually aggregate and reconcile concepts derived from the literature, manual coding, and model-assisted coding, by aligning and merging overlapping and redundant concepts into broader strategies.\footnote{For example, we initially obtained separate codes for utterances such as ``we need to solve correctly'' and ``but remember the correct answer is \ldots''. As both refer to the same underlying concept, namely explicitly referencing the correct answer (\texttt{CORR}), we merged them into a single category.}
This yields the taxonomy in \cref{tab:taxonomy-definition}.
All ten strategies in our taxonomy were initially identified via manual open coding. LLM-assisted coding independently recovered all strategies except \texttt{RECON} and proposed additional codes that were excluded post hoc, as they captured problem-specific rather than general strategies.
Eight of the ten final strategies are directly supported by distractor-generation literature (see Appendix~\cref{app:literature-tag-mapping}). The remaining two (\texttt{INTER} and \texttt{RECON}) reflect general reasoning behaviors: \texttt{INTER} reflects task interpretation under structured item constraints, and \texttt{RECON} reflects iterative backtracking during solution refinement \citep{mislevy2003ecd}.
Further details on strategy reconciliation are provided in Appendix~\cref{app:taxonomy-construction}.\looseness=-1

\subsection{Trace Annotation}
\label{sec:method-annotation}
To characterize the distribution of strategies LLMs use during distractor generation, we scale up annotation by applying the taxonomy to all traces, marking each span with a strategy tag as illustrated in \cref{fig:main-annotation-example}. Following prior work, we use an LLM as the primary annotator and validate the resulting annotations manually on a representative subset of traces~\citep{thoughtology2025}, comparing insertions and deletions against human-annotated references. The LLM is prompted with (i) a description of the taxonomy, (ii) a small set of hand-picked examples per strategy tag grounded in observations from manual open coding, (iii) one full trace, and (iv) an instruction to re-generate the trace with strategy tags inserted after each span, e.g., \texttt{\textless{}PLAUS\textgreater{}} after ``Yes, this is a plausible distractor''. Annotating an entire trace in a single pass yields poor results due to length, so we split each trace into chunks and annotate each chunk separately; to preserve problem-specific context across chunks, we augment each chunk's prompt with examples extracted from the full trace (see Appendix~\cref{app:method-annotation} for details). Experimental details and validation results are reported in \cref{sec:experiments-process-analysis}.\looseness=-1

\section{Experiments}
We first establish that LLM's reasoning strategies help with distractor generation (\cref{sec:experiments-performance}), then examine how LLMs use and orchestrate the strategies of our taxonomy (\cref{sec:experiments-process-analysis}), and finally discuss alignment with best practices and failure modes (\cref{sec:discussion}).\looseness=-1

\paragraph{Datasets.}
\label{sec:experimental-setup}
For both taxonomy construction and experiments, we use the Eedi Math~\citep{eedi-mining-misconceptions-in-mathematics} and SciQ~\citep{welbl-etal-2017-crowdsourcing} datasets, both MCQ datasets with three distractor options per item.
Eedi serves as our primary dataset, as it contains grade-school mathematics problems with annotated high-level student misconceptions and is widely used in prior work on distractor generation and misconception modeling~\citep{feng-etal-2024-exploring,lee2024mathmultiplechoicequestion,parikh-etal-2025-lookalike}.
We also include SciQ, a science QA dataset, to validate that our findings generalize across domains. While it does not come with misconception annotations, these are not required for taxonomy construction or our main trace-level analysis.
We remove image-based questions and questions that depend on answer choices (e.g., ``Which of the following is correct?''; see Appendix~\cref{app:preprocess-dataset} for further details).
For SciQ, we randomly sample 500 questions; for Eedi, we keep only MCQs with annotated misconceptions, yielding 429 problems.\looseness=-1

\paragraph{Models and prompting.}
We use two reasoning models, \ds \citep{deepseekai2025deepseekv32pushingfrontieropen} and \glm~\citep{5team2025glm45agenticreasoningcoding}, across the two datasets and three prompting settings: (i) \defn{direct prompting}, where reasoning is disabled and models directly output distractors;
(ii) \defn{chain-of-thought (CoT) prompting} \citep{wei2022cot}, where models are prompted to think step-by-step before generating distractors but reasoning is disabled; (iii) \defn{reasoning prompting}, where models generate additional reasoning tokens before outputting distractors. Only settings (ii) and (iii) produce intermediate traces that can be analyzed.
All settings share the same base \defn{neutral prompt} for distractor generation, with setting-specific adaptations described in Appendix~\cref{app:prompts-collect};
the LLM is prompted for each MCQ separately.

\paragraph{Performance metrics.}
To stratify traces by performance and evaluate the effect of reasoning on LLM distractor generation (\cref{sec:experiments-performance}), we define three evaluation metrics.
Each dataset $\{p_n\}_{n=1}^N$ contains $N$ problems $p_n=(x_n, y_n, \mathcal{Z}_n)$, where $x_n$ is the textual description of the problem, $y_n$ is the correct answer choice, and $\mathcal{Z}_n$ with $|\mathcal{Z}_n|=3$ is the set of (three) ground-truth distractors.
For the problem indexed by $n$, we sample candidate distractors from an LLM, yielding a multiset $\hat{Z}_n=(\hat{\mathcal{Z}}_n, m_n)$, where $\hat{\mathcal{Z}}_n$ is the underlying set and $m_n\colon \hat{\mathcal{Z}}_n \rightarrow \mathbb{N}$
is the multiplicity of the elements in $\hat{\mathcal{Z}}_n$ (i.e., their counts);
we write $|\hat{Z}_n| \defeq \sum_{z \in \hat{\mathcal{Z}}_n} m_n(z)$ for the total candidate count.
In practice, we restrict $|\hat{Z}_n|$ to $3$ via prompting and post-processing. We use three metrics: \defn{proportional match} $|\mathcal{Z}_n \cap \hat{\mathcal Z}_n|/|\mathcal{Z}_n|$ \citep{feng-etal-2024-exploring,fernandez-etal-2024-divert}, the number of candidates that equal the correct answer choice (\defn{\#correct}) $m_n(y_n)$ (max value 3, optimal value 0), and the number of redundant candidate choices (\defn{\#repetitions}) $|\hat{Z}_n| - |\hat{\mathcal Z}_n|$ (max value 2, optimal value 0). We report the average across all $N$ problem instances and confidence intervals (CIs) based on t-distributions.
These metrics require equivalence checks between distractors, which are arbitrary token sequences and not restricted to numerical values. Following prior work~\citep{ross2025learningmakemistakesmodeling}, we use an LLM as judge. \texttt{gpt-4-1-mini}~\citep{openai2024gpt4technicalreport} achieves a macro-average precision of 98\% and recall of 96\%, which we verified through manual validation of 1000 random comparisons. Further details are provided in Appendix~\cref{app:measure-performance}.\looseness=-1

\begin{table}
\centering

\small

\begin{tabular}{llcc}
\toprule
\textbf{Model} & \textbf{Setting} & \textbf{Eedi} & \textbf{SciQ} \\
\midrule
\multirow{3}{*}{\textbf{\dsabbr}}
& Direct     & 0.34 $\pm$ 0.03 & 0.08 $\pm$ 0.02 \\
& CoT        & 0.51 $\pm$ 0.03 & 0.13 $\pm$ 0.02 \\
& Reasoning  & 0.52 $\pm$ 0.03 & 0.14 $\pm$ 0.02 \\
\midrule
\multirow{3}{*}{\textbf{\glm}}
& Direct     & 0.42 $\pm$ 0.03 & 0.12 $\pm$ 0.02 \\
& CoT        & 0.50 $\pm$ 0.03 & 0.15 $\pm$ 0.02 \\
& Reasoning  & 0.52 $\pm$ 0.03 & 0.15 $\pm$ 0.02 \\
\bottomrule
\end{tabular}
\caption{Proportional match (mean ± 95\% CI based on t-distribution) for \ds (\dsabbr) and \glm, on the Eedi and SciQ datasets, under different prompting settings.
\looseness=-1
}
\label{tab:distr-performance}
\vspace{-5pt}
\end{table}

\subsection{Reasoning Improves Performance}
\label{sec:experiments-performance}
Before examining reasoning traces, we first assess what output-level performance metrics reveal about the usefulness of reasoning for LLMs' distractor generation.
We find that proportional match improves significantly when moving from direct prompting to either CoT or reasoning for both LLMs and datasets (\cref{tab:distr-performance}). Furthermore, as detailed in Appendix~\cref{app:full-results-performance}, on Eedi \#correct decreases by $\approx 92\%$ and \#repetitions by $82\%$; on SciQ \#correct drops by $82\%$ while \#repetitions remain low overall.
These patterns suggest that the additional tokens contain useful distractor-generation strategies, with the drops in \#correct and \#repetitions pointing to successful discrimination checks and curation.
We interpret the substantially higher proportional match on Eedi than on SciQ as partly reflecting task structure: misconception-based construction may constrain the output space to specific plausible errors, making alignment with human-authored distractors more tractable.

\subsection{The LLM Distractor Generation Process}
\label{sec:experiments-process-analysis}
We present three complementary views of LLMs' distractor-generation processes: overall strategy usage, how strategy usage evolves throughout a trace, and how strategies are organized into reasoning subprocesses.
We report \ds with reasoning enabled; reported trends are consistent across \glm and the CoT setting, with full results in Appendix~\cref{app:full-results-occurrences-of-strategies,app:full-results-strategies-over-time,app:full-results-common-sequences-of-strategies}.\looseness=-1

\paragraph{Annotation.}
We apply the pipeline from \cref{sec:method-annotation} to 885 traces, stratified by dataset, model, CoT versus reasoning, and performance (proportional match above vs.\ below 0.5). We validate annotations on a representative subset of 64 traces using \gptfourone as annotator, achieving a precision of 0.97 and recall of 0.95, macro-averaged across strategy tags, datasets, and models; a detailed breakdown is in Appendix~\cref{app:method-annotation}.\looseness=-1

\paragraph{Occurrences of strategies.}
We analyze the average number of annotated spans per trace for each strategy of our taxonomy in \cref{tab:component_presence}. \texttt{INST} is among the most frequently observed strategies across both datasets,\footnote{This is expected, given that repeated occurrences are counted separately and statements involving \texttt{PLAUS}, \texttt{DISCR} and \texttt{CURATE} often reference \texttt{INST}.} and \texttt{INTER}, \texttt{CORR}, \texttt{PLAUS}, and \texttt{CURATE} also occur consistently. By contrast, \texttt{LINK} is prevalent on SciQ but nearly absent on Eedi, whereas \texttt{ERR\_DESC} and \texttt{ERR\_SIM} dominate on Eedi but appear only sporadically on SciQ. \texttt{DISCR} and \texttt{RECON} show smaller but consistent differences across datasets: \texttt{DISCR} is more frequent on SciQ and \texttt{RECON} on Eedi.
The consistent presence of \texttt{INTER}, \texttt{CORR}, \texttt{INST}, \texttt{PLAUS}, and \texttt{CURATE} across both datasets suggests a shared high-level structure involving problem understanding, candidate construction, evaluation and selection. At the same time, the dataset-specific patterns are consistent with two distinct approaches to candidate construction and evaluation: misconception-based on Eedi, where \texttt{ERR\_DESC} and \texttt{ERR\_SIM} drive candidate generation, and similarity-based on SciQ, where candidates are identified via \texttt{LINK} and their incorrectness is verified via \texttt{DISCR}.\looseness=-1

\begin{table}
\centering

\small
\begin{tabular}{l r@{}c@{}l r@{}c@{}l}
\toprule
\textbf{Strategy} & & \makebox[0pt]{\textbf{Eedi}} & & & \makebox[0pt]{\textbf{SciQ}} & \\
\midrule
Task Interpretation    & 11.22 & $\;\pm\;$ & 1.38 & 5.84  & $\;\pm\;$ & 0.80 \\
Correct Answer Ref.    & 9.33  & $\;\pm\;$ & 1.40 & 5.52  & $\;\pm\;$ & 0.89 \\
Conceptual Link        & 0.04  & $\;\pm\;$ & 0.05 & 5.36  & $\;\pm\;$ & 0.87 \\
Error Description      & 22.83 & $\;\pm\;$ & 2.70 & 0.78  & $\;\pm\;$ & 0.31 \\
Error Simulation       & 5.70  & $\;\pm\;$ & 1.18 & 0.00  & $\;\pm\;$ & 0.00 \\
Outcome Instantiation  & 34.84 & $\;\pm\;$ & 4.05 & 24.71 & $\;\pm\;$ & 2.95 \\
Plausibility Check     & 7.94  & $\;\pm\;$ & 1.15 & 5.33  & $\;\pm\;$ & 0.90 \\
Discriminability Check & 2.54  & $\;\pm\;$ & 0.59 & 10.34 & $\;\pm\;$ & 1.50 \\
Final Set Curation     & 3.34  & $\;\pm\;$ & 0.39 & 5.83  & $\;\pm\;$ & 0.60 \\
Reconsideration        & 5.46  & $\;\pm\;$ & 0.71 & 1.05  & $\;\pm\;$ & 0.23 \\
\bottomrule
\end{tabular}
\caption{Average occurrences of taxonomy strategies (\cref{tab:taxonomy-definition}) in reasoning traces for \ds (mean ± 95\% CI based on t-distribution) on both datasets.\looseness=-1
}
\label{tab:component_presence}

\vspace{-5pt}
\end{table}

\paragraph{Strategies over time.}

\begin{figure}[!t]

    \centering
    \includegraphics[width=1.0\linewidth, trim=5 5 5 5, clip]{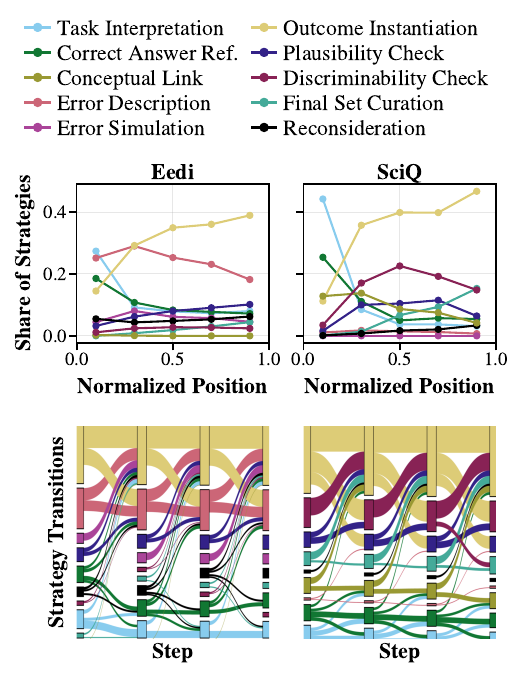}

    \caption{Strategy usage in \ds reasoning traces on Eedi, shown in two views (strategies defined in \cref{tab:taxonomy-definition}). Top: temporal distribution across five normalized stages over the span of a reasoning trace (0 = start, 1 = end), where proportions sum to one within each bin. Bottom: transition structure between strategies over sequences of length 4 (left to right), where node height reflects strategy prevalence and edge width encodes transition probability. Only dominant transitions ($>$15\% outgoing probability) are shown for clarity.\looseness=-1}

    \label{fig:transition-probabilities}
    \label{fig:components-over-time}

    \vspace{-5pt}
\end{figure}

We track the proportion of each strategy across five temporal bins, where temporal position within each trace is normalized to $[0, 1]$ by character offset.
The resulting graphs in the top row of \cref{fig:components-over-time} reveal clear shifts in strategy usage throughout the reasoning process.
Early reasoning is usually dominated by task interpretation (\texttt{INTER}) and identifying the correct answer (\texttt{CORR}), while usage of final set curation (\texttt{CURATE}) increases towards the end.
For Eedi, error description (\texttt{ERR\_DESC}) peaks right after \texttt{INTER} and \texttt{CORR}, also coinciding with an increase in instantiation of incorrect outcomes (\texttt{INST}). Toward the end, plausibility assessment (\texttt{PLAUS}) becomes more frequent.
For SciQ, conceptual linking (\texttt{LINK}) appears more frequently in the beginning, while plausibility (\texttt{PLAUS}) and discriminability checks (\texttt{DISCR}) are performed across the middle and later stages.
These temporal profiles are consistent with a progression from problem understanding to candidate construction to evaluation and selection.
We observe similar trends for CoT but with \texttt{INTER} and \texttt{CORR} taking up a larger share of the reasoning process, consistent with fewer distractor candidates being explored (see Appendix~\cref{app:full-results-strategies-over-time}).

\paragraph{Common sequences of strategies.}

To analyze the typical sequences in which strategies appear, we examine transition probabilities between strategies within sliding windows of 4 consecutive strategy annotations, shown in the bottom row of \cref{fig:transition-probabilities}.
The transitions reveal functional groupings. On Eedi, \texttt{ERR\_DESC} and \texttt{INST} are tightly coupled. From the traces we examined during open coding and annotation validation, we observe two recurring sub-patterns: models either list multiple error descriptions before instantiating them, or iteratively interleave the two, sometimes with \texttt{PLAUS}. On SciQ, the coupled trio is \texttt{INST}, \texttt{PLAUS}, and \texttt{DISCR}, reflecting a quality-control focus in the construction phase. Self-transitions for \texttt{INTER} and \texttt{CORR} indicate that these problem-understanding strategies tend to extend across multiple consecutive annotation spans.
For CoT on Eedi (see Appendix~\cref{app:full-results-common-sequences-of-strategies}), a common \texttt{ERR\_DESC} $\rightarrow$ \texttt{ERR\_SIM} $\rightarrow$ \texttt{INST} transition suggests
that LLMs explicitly simulate the erroneous reasoning step, possibly ensuring consistency between the error description and its instantiation.

\subsection{Discussion and Implications}
\label{sec:discussion}
\paragraph{Shared strategy structure and domain adaptation.} Across models, datasets, and prompts, the analyses above converge on a shared high-level distractor-generation pipeline: problem understanding, candidate construction, evaluation and selection. This pipeline is exploratory: after anchoring to the correct answer, models over-generate candidate distractors and then select a final set, resembling a best-of-$N$ search. \ds in reasoning mode considers, on average, $12.6$ alternative incorrect solution paths on Eedi and $9.9$ on SciQ before final selection, compared to $5.9$ and $5.5$ under CoT prompting. These counts are based on manual analysis of 16 randomly selected problems per setting.
What differs across domains is the candidate construction and evaluation stage. We hypothesize this reflects task structure: on Eedi, where distractors are anchored to annotated student misconceptions, describing and instantiating these errors (\texttt{ERR\_DESC}, \texttt{ERR\_SIM}, \texttt{INST}) mostly yields already-incorrect candidates, reducing the need for explicit discriminability checks. On SciQ, models instead use \texttt{LINK} to construct candidates via semantic relatedness to the correct answer and \texttt{DISCR} to verify they are indeed incorrect, consistent with \texttt{LINK} appearing early in the trace and \texttt{DISCR} later.

\paragraph{Comparison with literature best practices.}
On Eedi, where misconception-based construction is both applicable and the learning-science-preferred approach, we can directly test whether LLMs actually follow this approach. The strategy distributions and temporal profiles indicate that they predominantly do, following a solve-first, error-injection process in which a potential student error is articulated, instantiated through erroneous reasoning, and used to produce a distractor, rather than relying on surface-level similarity heuristics.
A targeted trace-level analysis further confirms this (see Appendix~\cref{app:solve-first-error-inject}): models construct a step-by-step solution first in 95.2\% of traces on average across models, and among these solve-first traces, diverge at a specific step to inject an error in 73.8\% of cases.
Conversely, similarity-based strategies are rare ($\approx$3\% of traces) and largely confined to narrow settings such as rounding or magnitude-estimation problems.
Beyond this structural alignment, we also examine the cognitive stance models adopt when reasoning about errors: when prompted to simulate a specific student error, 99.3\% of \ds traces reason from an expert perspective (e.g., ``the student will likely do\ldots'' or ``I must determine what the student would likely do\ldots'') rather than adopting a student's first-person stance (see Appendix~\cref{app:error-simulation}).

\paragraph{Diagnosing failure modes.}
We analyze multiple stages of LLMs' distractor-generation pipeline to determine whether failures stem from (i) incorrect error simulation, (ii) the absence of an effective reasoning structure, (iii) failure to recover the correct solution, or (iv) inaccurate plausibility assessment and candidate selection. We find that the first two rarely fail, whereas the latter two are the dominant bottlenecks, as detailed below.

\textbf{(i) Error simulation rarely fails.}
When prompted to simulate a problem-specific student error (sampled from Eedi's misconception annotations for each of the 429 problems), \ds achieves a mean accuracy of $0.92 \pm 0.02$, manually verified against \emph{any} faithful interpretation of the error (procedure in Appendix~\cref{app:error-simulation}). However, this does not guarantee that the model selects the most plausible interpretation: only $62\%$ of distractors judged faithful to a given error type appear in the human-authored ground-truth set, suggesting multiple valid instantiations exist.
Where the targeted distractor was not produced, the underlying problems tend to have longer correct solutions ($p<0.05$, Mann-Whitney U test, comparing solution step counts), suggesting the model is less likely to reproduce the targeted distractor on problems with longer solutions.

\textbf{(ii) Reasoning structure rarely fails.} A literature-informed prompt enforcing explicit multi-step reasoning (solving, enumerating errors, simulating them, evaluating plausibility, and selecting distractors; prompt in Appendix~\cref{app:literature-informed}) yields no significant improvement over the neutral prompt for either model (\ds: $0.52 \pm 0.03 \rightarrow 0.55 \pm 0.02$; \glm: $0.52 \pm 0.03 \rightarrow 0.52 \pm 0.03$; paired $t$-test, $p > 0.05$), suggesting that both models already self-organize an effective reasoning structure without explicit procedural guidance. This observation largely generalizes across seven further reasoning models (Appendix~\cref{app:model-diversity}).\footnote{The exception is the OpenAI models, for which enforcing the structure improves performance: \gptoss, \gptosslarge~\citep{openai2025gptoss120bgptoss20bmodel}, and \gptfive~\citep{singh2026openaigpt5card} benefit, whereas \gemma~\citep{gemmateam2025gemma3technicalreport}, \geminis~\citep{comanici2025gemini25pushingfrontier}, and \glmflash~\citep{5team2025glm45agenticreasoningcoding} do not. As the OpenAI models are also the lowest-scoring under the neutral prompt, model family and performance headroom are confounded.}

\textbf{(iii) Correct-solution recovery is a genuine bottleneck.} Providing the correct answer to the MCQ in the prompt improves proportional match on both Eedi ($0.52 \pm 0.03 \rightarrow 0.56 \pm 0.03$) and SciQ ($0.14 \pm 0.02 \rightarrow 0.18 \pm 0.03$; see footnote~\cref{fn:sciq-corr-improve-caveat}), confirming that unreliable solution recovery propagates through the pipeline and degrades downstream generation.

\textbf{(iv) Candidate selection is a second bottleneck.} We can quantify how much the model loses at selection by asking how many human-authored distractors it proposes somewhere in its reasoning but leaves out of its final three. Proportional match already counts how many reach the final set. To count how many the model proposes at any point instead, we introduce an \emph{oracle} ranker that matches each annotated \texttt{INST} candidate against the human-authored set (see Appendix~\cref{app:oracle-ranker}); the gap between the two is the selection loss.
The oracle raises \ds's proportional match from $0.52$ to $0.67$ on Eedi and from $0.14$ to $0.29$ on SciQ (paired $t$-test, $p<10^{-7}$), recovering about a third of the otherwise-missed distractors on Eedi and a sixth on SciQ. This suggests the model frequently proposes a valid distractor and then discards it.
Part of this loss might be attributable to the model's own plausibility and discriminability checks (\texttt{PLAUS}, \texttt{DISCR}): resolving each check to the candidate it evaluates and classifying whether it argues for or against keeping it (details in Appendix~\cref{app:self-check-drops}), we find that the model argues against keeping $28\%$ of the candidates that match human-authored ones on Eedi and $15\%$ on SciQ under \ds (the same pattern holds for \glm). The checks do discriminate, rejecting invalid candidates $1.5$--$1.9\times$ as often as a valid ones, but they still turn a non-trivial share of valid candidates negative. Even a perfect ranker cannot recover a distractor the model never proposed, and $33\%$ of human-authored distractors on Eedi and $71\%$ on SciQ never appear in the reasoning at all. These misses point to a failure of candidate \emph{generation} rather than selection, especially on SciQ.

\textbf{Characteristics of high- and low-scoring traces.}
The stage analyses above pool all traces. We can instead split individual traces by output quality into high- and low-scoring at a proportional-match threshold of $0.5$. On Eedi the strategy profiles differ between the two groups but not on SciQ, and the difference on Eedi is concentrated in one strategy: low-scoring traces enumerate student errors (\texttt{ERR\_DESC}) more. This is significant under reasoning and borderline under CoT.\footnote{Significance comes from permutation tests: PERMANOVA~\citep{anderson2001permanova,mcardle2001} for whether the overall ten-strategy profile differs, then one test per strategy with the significance threshold tightened for running ten of them (see Appendix~\cref{app:behaviour-performance}).} Low-scoring traces are longer and may come from harder problems, but the gap is neither a length nor a difficulty effect: it holds as a share of strategies and under grouped solution-step counts. Two strategies, correct-answer reference (\texttt{CORR}) and task interpretation (\texttt{INTER}), have a higher \emph{share} in high-scoring traces but no higher count: these traces contain fewer tags overall, so a similar count forms a larger fraction, not more of either strategy (\cref{app:behaviour-performance}). The error-enumeration gap itself may have a mechanical source: problems with a larger space of plausible errors may draw more enumeration and make any single human-authored distractor statistically harder to match.

\section{Conclusion}
We introduced a taxonomy of ten distractor-generation strategies grounded in learning-science principles and observed model behavior, and applied it to reasoning traces from two LLMs across math and science MCQs.

Across all settings, we find a shared high-level structure involving problem understanding, anchoring to the correct solution, exploratory candidate construction, and iterative evaluation and selection, resembling a best-of-$N$ search over incorrect solution paths. On Eedi, where all human-annotated distractors reflect specific misconceptions, models follow a solve-first, error-injection approach, constructing a step-by-step solution in 95.2\% of traces and diverging to inject an error in 73.8\% of those cases. This is in close alignment with the preferred misconception-based approach in the educational literature. However, on SciQ, they instead rely more on semantic similarity and explicit discriminability checks.

Not all parts of this distractor-generation pipeline are equally reliable, however: error simulation and the implicit organization of multi-step reasoning rarely fail, while failures concentrate in correct-solution recovery and downstream candidate selection. Providing the correct solution in the prompt yields significant relative improvements of 6.4\% on Eedi and 30.7\% on SciQ (see \cref{fn:sciq-corr-improve-caveat}), confirming correct-solution recovery as a primary bottleneck.

Together, these findings offer a concrete answer to how LLMs arrive at incorrect-but-plausible reasoning: through a structured but domain-adaptive process that mirrors expert practice on math problems while exposing clear points of failure across both domains. By isolating these failure points, our analysis points to targeted interventions, with implications for automated assessment, intelligent tutoring systems, and scalable creation of high-quality educational content.

\section*{Limitations}
\label{sec:limitations}

Our analysis is subject to several limitations. First, our analysis characterizes \emph{what the reasoning trace says}, not necessarily \emph{what the model internally computes}. Reasoning traces are not guaranteed to faithfully reflect the underlying decision process, so the observed strategies should be interpreted as descriptions of externalized reasoning rather than mechanistic claims.
Prior work argues, however, that reasoning traces are structured and shape model outputs rather than being mere post-hoc narration \citep{thoughtology2025}.

Second, both studied models are open-weight reasoning models. Reasoning models are the natural starting point for process-level analysis, as they produce explicit intermediate traces that can be directly annotated with strategies; however, non-reasoning LLMs and closed-source models may exhibit different strategy distributions, and our quantitative findings should not be extrapolated to those settings without further study.

Finally, performance is measured against a finite set of human-annotated distractors, which does not account for further plausible distractors that are absent (see \cref{sec:discussion}). On SciQ, this concern is more acute: distractors were not curated to reflect specific student misconceptions, making proportional match a weaker benchmark signal on that dataset. Both issues primarily affect proportional match as a benchmark metric; our core contribution is the process-level analysis, which does not depend on ground truth completeness.

\section*{Acknowledgments}
We thank Eedi and the creators of SciQ for making their datasets publicly available.

\bibliography{custom}

\clearpage

\appendix

\tcbset{
  promptstyle/.style={
    breakable, enhanced jigsaw,
    lines before break=5,
    colback=white, colframe=black!55,
    boxrule=0.4pt, arc=1pt, outer arc=1pt,
    left=5pt, right=5pt, top=4pt, bottom=4pt, boxsep=0pt,
    parbox=false,
    before skip=\medskipamount, after skip=3pt,
  },
}
\newenvironment{promptbox}
  {\begin{tcolorbox}[promptstyle]
   \scriptsize\raggedright
   \clubpenalty=10000 \widowpenalty=10000
   \setlength{\parindent}{0pt}\setlength{\parskip}{0pt}}
  {\end{tcolorbox}}
\newcommand{\promptrule}{\tcbline}

\newcommand{\promptcaption}[2]{
  \par\addvspace{3pt}
  \noindent\begin{minipage}{\linewidth}
    \captionsetup{hypcap=false}
    \captionof{table}{#1}\label{#2}
  \end{minipage}\par
  \addvspace{\medskipamount}}

\clearpage

\section{Supplementary Material - Experimental Setup}

\subsection{Preprocessing the Dataset}
\label{app:preprocess-dataset}
\paragraph{Removing problems relying on choices.}
Both the Eedi and SciQ datasets contain problems of the form ``Which of the following is correct?'', which rely on the provided answer choices to be solvable. This is problematic for distractor generation because it implies an effectively infinite answer space. Therefore, we manually reviewed all problems in the dataset to determine whether they admit a unique solution without the provided choices, and removed those that do not.

\subsection{Measuring Performance}
\label{app:measure-performance}
\paragraph{Equivalence checks.}
The metrics defined in \cref{sec:experimental-setup} require checking whether candidate distractors $\hat{\mathcal Z}$ are equivalent to the annotated distractors $\mathcal Z$. Requiring exact matches, while eliminating false positives, would be far too strict:
most problems' answers may be expressed in multiple ways (e.g., $\frac{1}{2}$ vs. $1/2$) or equivalence may depend on the problem description (e.g., $\frac{1}{2}$ is not equivalent to $0.5$ if the problem asks for it to be converted to decimal).
Following prior work, we use an LLM (\gptfouronemini~\citep{openai2024gpt4technicalreport}, prompted to judge equivalence of two provided answers) to judge semantic equivalence when exact matches come back negative \citep{ross2025learningmakemistakesmodeling}.
Our policy is to enforce output-formatting constraints that a problem states explicitly, but not to infer constraints that are left unstated. 
For instance, if a problem explicitly requires its answer as a fully reduced expression, $\frac{6}{3}$ and $2$ are treated as non-equivalent, whereas if it imposes no such requirement they are considered equivalent. This mirrors how the ground-truth distractors themselves are written and avoids penalizing candidates for surface form when the problem does not constrain it.
Prompts are split into versions with examples specific to Eedi (\cref{tab:equivalence_prompt_eedi}) and to SciQ (\cref{tab:equivalence_prompt_sciq}).
The LLM achieves 98\% precision and 92\% recall in equivalence judgements over 500 Eedi comparisons and 97\% precision and 100\% recall over 500 SciQ comparisons.

\begin{promptbox}
\textbf{System Prompt} \par
\promptrule
{
\ttfamily
You are an AI assistant tasked with judging whether two answer choices to a middle-school multiple-choice math problem are semantically the same as one another. You must not solve the problem and not evaluate factual correctness --- only compare the two answers with one another relative to the problem's formatting requirements.

Your output must follow this exact structure:

\texttt{<format> $[$TRUE/FALSE$]$ </format>} \par
\texttt{<equivalent> $[$TRUE/FALSE$]$ </equivalent>}

Meaning of \texttt{<format>}:

Output TRUE if the problem explicitly requires a specific numeric format, such as:

-- rounding to a given number of decimal places or significant digits \par
-- expressing the answer in scientific notation \par
-- expressing the answer as a simplified fraction \par
-- expressing the answer in terms of a constant (e.g., ``in terms of $\pi$'') \par
-- any other explicitly stated formatting requirement

Ignore unit requirements (e.g., ``in cm'' does NOT count as a specific format).

Output FALSE if the problem does not explicitly specify a numeric format.

Meaning of \texttt{<equivalent>}:

Determine whether \texttt{answer\_1} and \texttt{answer\_2} represent the same value or concept under the rules:

If \texttt{<format>} is FALSE (no required numeric format):

Two answers are equivalent if:

-- they have the same mathematical value (e.g., $3.1 = 31/10$) \par
-- they differ only in non-semantic aspects (e.g., LaTeX wrappers, capitalization, filler words)

Examples of equivalent under \texttt{<format>} = FALSE:

-- 10, 10.0, $\backslash(10\backslash)$ \par
-- 3.1, 31/10 \par
-- fourteen, 14 \par
-- Only Bob, Bob

If \texttt{<format>} is TRUE (specific format required):

Two answers are equivalent only if:

1. They represent the same mathematical value, AND \par
2. They are both expressed in the required format.

This means:

-- 3.14 vs.\ 3.140 (when rounding to 2 decimal places required) $\rightarrow$ not equivalent \par
-- $4\pi$ vs.\ 12.56 (when ``in terms of $\pi$'' required) $\rightarrow$ not equivalent \par
-- 3.1 vs.\ 31/10 (when ``round to one decimal place'' required) $\rightarrow$ not equivalent

Ignore trivial formatting wrappers (e.g., $31/10 = \backslash(31/10\backslash)$).

General Rules:

-- Do not solve the problem. \par
-- Do not judge correctness of the answers. \par
-- Only compare \texttt{answer\_1} with \texttt{answer\_2}. \par
-- \texttt{answer\_1} and \texttt{answer\_2} can be equivalent regardless of whether they are correct or not.
\par}
\promptrule
\textbf{User Prompt} \par
\promptrule
{
\ttfamily
\texttt{<problem>} \promptvar{\{problem\_formulation\}} \texttt{</problem>} \par
\texttt{<answer\_1>} \promptvar{\{answer\_a\}} \texttt{</answer\_1>} \par
\texttt{<answer\_2>} \promptvar{\{answer\_b\}} \texttt{</answer\_2>}
\par}
\end{promptbox}
\promptcaption{Prompt for judging equivalence between the two answers \promptvar{\{answer\_a\}} and \promptvar{\{answer\_b\}} to a math multiple-choice question \promptvar{\{problem\_formulation\}} of Eedi.}{tab:equivalence_prompt_eedi}

\begin{promptbox}
  \textbf{System Prompt} \par
\promptrule
  {
  \ttfamily
  You are an AI assistant tasked with judging whether two answer choices to a middle-school multiple-choice science question are semantically the same as one another. You must not answer the
  question and not evaluate factual correctness --- only compare the two answers with one another relative to what the question is asking.

  Your output must follow this exact structure:

  \texttt{<equivalent> $[$TRUE/FALSE$]$ </equivalent>}

  Meaning of \texttt{<equivalent>}:

  Two answers are equivalent if they refer to exactly the same scientific concept, entity, term, or value, differing only in:

  -- capitalization, punctuation, pluralization, articles, or filler words \par
  -- trivial formatting wrappers (e.g., LaTeX, extra whitespace) \par
  -- abbreviation vs.\ its expansion (e.g., DNA vs.\ deoxyribonucleic acid) \par
  -- well-established synonyms denoting the exact same referent (e.g., heat vs.\ thermal energy) \par
  -- rewording that preserves the exact same referent without adding or removing entities

  Examples of equivalent:

  -- photosynthesis = Photosynthesis \par
  -- planet = a planet \par
  -- mirrors = a mirror \par
  -- DNA = deoxyribonucleic acid \par
  -- heat = thermal energy \par
  -- products = the products

  Two answers are NOT equivalent if:

  -- they name different scientific concepts, even if closely related (e.g., photosynthesis vs.\ respiration, mirrors vs.\ lenses, circadian rhythm vs.\ circannual rhythm) \par
  -- one is a subset or superset of the other, or introduces or removes entities (e.g., ``protons and neutrons'' vs.\ ``protons'') \par
  -- they give different specific values or quantities (e.g., ``less than half'' vs.\ ``5\%'', ``type 1'' vs.\ ``type 2 diabetes'') \par
  -- they specify different levels of generality that change the referent (e.g., ``bacteria'' vs.\ ``E.\ coli'')

  Examples of NOT equivalent:

  -- mirrors $\neq$ lenses \par
  -- protons and neutrons $\neq$ protons \par
  -- carbon $\neq$ nitrogen \par
  -- heat $\neq$ ultraviolet

  General Rules:

  -- Do not answer the question. \par
  -- Do not judge correctness of the answers. \par
  -- Only compare \texttt{answer\_1} with \texttt{answer\_2}. \par
  -- \texttt{answer\_1} and \texttt{answer\_2} can be equivalent regardless of whether they are correct or not.
  \par}
\promptrule
  \textbf{User Prompt} \par
\promptrule
  {
  \ttfamily
  \texttt{<problem>} \promptvar{\{problem\_formulation\}} \texttt{</problem>} \par
  \texttt{<answer\_1>} \promptvar{\{answer\_a\}} \texttt{</answer\_1>} \par
  \texttt{<answer\_2>} \promptvar{\{answer\_b\}} \texttt{</answer\_2>}
  \par}
\end{promptbox}
\promptcaption{Prompt for judging equivalence between two answer choices \promptvar{\{answer\_a\}} and \promptvar{\{answer\_b\}} to a middle-school multiple-choice science question
  \promptvar{\{problem\_formulation\}} of SciQ.}{tab:equivalence_prompt_sciq}

\subsection{Collecting Reasoning Traces and Evaluating Answer Equivalence}
\label{app:prompts-collect}
We provide the prompt used for distractor generation in both the direct and reasoning settings (\cref{tab:direct_distractor_prompt}), as well as the chain-of-thought prompt that enables the non-reasoning models to generate distractors step-by-step (\cref{tab:cot_distractor_prompt}).
For all models, we set token limits to 16k for reasoning and 8k for CoT.

\begin{promptbox}
\textbf{System Prompt} \par
\promptrule
{
\ttfamily
You will be given a \textcolor{purple}{\{math,science\}} question \textcolor{blue}{along with the correct answer}. Please generate 3 incorrect distractor answers for the question to be used as multiple-choice options in a multiple-choice exam. \par

$[$Template$]$ \par
Distractor1: \par
\ldots \par
Distractor3:
\par}
\promptrule
\textbf{User Prompt} \par
\promptrule
{
\ttfamily
Question: \promptvar{\{problem\_formulation\}}\par
\textcolor{blue}{Answer: \promptvar{\{correct\_answer\}}}
\par}
\end{promptbox}
\promptcaption{Neutral prompt for generating distractors for one math problem \promptvar{\{problem\_formulation\}}. Parts in \textcolor{blue}{blue} are shown only when the correct answer \promptvar{\{correct\_answer\}} is revealed. We use \textcolor{purple}{math} for Eedi and \textcolor{purple}{science} for SciQ.}{tab:direct_distractor_prompt}

\begin{promptbox}
\textbf{System Prompt} \par
\promptrule
{
\ttfamily
You will be given a \textcolor{purple}{\{math,science\}} question. Please generate 3 incorrect distractor answers for the question to be used as multiple-choice options in a multiple-choice exam. Think step-by-step before giving your final answer. Output only your step-by-step reasoning and the final distractors like so:

$[$Step-By-Step$]$\par
Let's think step-by-step, ...\par

$[$Final Answer$]$ \par
Distractor1: \par
\ldots \par
Distractor3:
\par}
\promptrule
\textbf{User Prompt} \par
\promptrule
{
\ttfamily
Question: \promptvar{\{problem\_formulation\}}
\par}

\end{promptbox}
\promptcaption{Neutral chain-of-thought (CoT) prompt for generating distractors for one math problem \promptvar{\{problem\_formulation\}}. We use \textcolor{purple}{math} for Eedi and \textcolor{purple}{science} for SciQ.}{tab:cot_distractor_prompt}
\subsection{Details on Taxonomy Construction}
\label{app:taxonomy-construction}
During taxonomy construction, several additional and potentially complementary strategies were identified through both human and LLM-based open coding but were ultimately excluded. Table~\ref{tab:discarded-strategies} summarizes these candidates, gives illustrative examples, their origin, and the rationale for exclusion.\looseness=-1

\begin{table*}[!htbp]
\centering
\small
\setlength{\tabcolsep}{4pt}
\begin{tabular}{p{0.2\textwidth} p{0.30\textwidth} p{0.12\textwidth} p{0.33\textwidth}}
\toprule
\textbf{Strategy} & \textbf{Example / Description} & \textbf{Identified By} & \textbf{Reason for Exclusion} \\
\midrule

\textbf{Factual Recall} & Explicit recall of relevant background information & Human & Only observed as part of \texttt{CORR}. \\

\textbf{Diagnostic Value} & Stating the diagnostic value of a distractor (e.g., ``tests the student's attention to concept x'') & Human & Implicitly captured via \texttt{ERR\_DESC} and \texttt{LINK}. \\

\textbf{Properties of Good Distractors} & Listing properties of good distractors. & Human & Only observed as part of \texttt{INTER}, \texttt{LINK} and \texttt{CURATE}. \\

\textbf{Response Formatting} & ``We need to produce them in the format Distractor1:\ldots'' & Both & Reasoning about output format required by our specific prompt rather than a general reasoning strategy. \\

\textbf{Recall Task-Specific Knowledge} & ``First, let's recall the basic Roman numeral symbols \ldots'' & Both & Explicitly observed only once and typically occurred implicitly within \texttt{CORR} or \texttt{ERR\_DESC}. \\

\textbf{Leveraging Domain-Specific Constraints} & Enumerating limited answer space (e.g., ``Who out of the two is correct?''). & Both & Problem-specific enumeration strategy; excluded to maintain focus on general reasoning strategies. \\

\textbf{Exploration of Unusual Distractors} & Claim that highly unusual or creative distractors were intentionally generated (no citation provided). & LLM & Conceptually overlaps with \texttt{ERR\_DESC} and \texttt{PLAUS}; insufficiently well-defined as a standalone strategy. \\

\textbf{Distractor Source Diversification} & Explicit effort to ensure distractors differ in type or origin to avoid similarity. & LLM & Strong overlap with \texttt{CURATE}. \\

\textbf{Verification and Self-Correction} & Revisiting earlier step-by-step reasoning towards distractors, potentially correcting issues. & LLM & Overlaps with \texttt{RECON} and often difficult to distinguish from \texttt{ERR\_SIM}. \\

\textbf{Detailed Error Description} & Captures specific behaviors such as incorrectly converting fractions to decimals without broader strategic framing. & LLM & Overlaps with \texttt{ERR\_SIM}. \\

\bottomrule
\end{tabular}
\caption{Discarded candidate strategies during the open coding phase of the taxonomy construction. Note that both the manual open coding as well as the LLM surfaced many additional strategies that were strictly captured by existing strategies; we only list the ones that might capture complementary aspects compared to the strategies they were merged into.
}
\label{tab:discarded-strategies}
\end{table*}

\paragraph{Merging strategies and coverage.}
Although the final taxonomy does not mirror all strategies that surfaced during open coding exactly, conceptual correspondence between strategies was always clear when merging them. For example, the LLM proposed the strategy ``Anchor on Correct Solution'' which we merged into \texttt{CORR}. In cases where titles differed, example citations from the traces frequently helped clarify the mapping. For instance, the proposed strategy ``Contextualize to Multiple-Choice Format'' with citation from the reasoning trace:

\begin{quote}
``Usually, the options are formatted similarly to the correct answer. Since correct is 0.256, options like 2.56, 0.0256, and 1.12 are all in decimal form with a few digits.''
\end{quote}
is naturally covered by \texttt{CORR} and \texttt{CURATE}.

\paragraph{Granularity of strategies.}
A key design decision in constructing the taxonomy concerns the level of granularity at which strategies are defined. For instance, \texttt{PLAUS} could be subdivided into more specific assessments, such as evaluating how common an error type is or identifying which constraints a distractor's derivation violates.
Although such finer-grained distinctions may offer additional analytical resolution, our objective is to capture \emph{general} strategy patterns rather than highly specialized variants. Increasing granularity would also likely reduce annotation reliability, as additional categories introduce greater complexity and less sharply defined decision boundaries.

\paragraph{Automatic open coding.}
We show the full prompt used to let \ds do open coding in \cref{tab:prompt-open-coding}.

\paragraph{Overview of literature sources of final taxonomy.}
\label{app:literature-tag-mapping}
The identified strategies largely align with prior literature. \texttt{LINK}, \texttt{PLAUS}, \texttt{DISCR}, and \texttt{CURATE} reflect item-writing guidelines that ensure distractors are semantically related to the correct answer, individually plausible, clearly incorrect, and collectively well-formed~\citep{haladyna2002,haladyna2013developing,gierl2017distractorreview,chamberlain2020integrity,ludewig2023features}.
\texttt{CORR}, \texttt{ERR\_DESC}, \texttt{ERR\_SIM}, and \texttt{INST} correspond to established distractor generation mechanisms, primarily misconception-based approaches~\citep{delatorre2009mcdina,ashlock2010error,shin2019multiple,tom2013mathquizes,gierl2017distractorreview}, and are further supported by repair theory and accounts of systematic student errors~\citep{brown1980repair,vanlehn1990mind}.
Finally, \texttt{INTER} and \texttt{RECON} capture general reasoning behaviors rather than distractor-specific mechanisms: \texttt{INTER} reflects interpretation of task requirements and output constraints under structured assessment design (e.g., Evidence-Centered Design)~\citep{mislevy2003ecd}, while \texttt{RECON} reflects iterative backtracking and revision during solution generation.

\begin{promptbox}
\textbf{System Prompt} \par
\promptrule
{
\ttfamily
You are an expert in cognitive task analysis, think-aloud protocol analysis, and grounded theory coding. Your task is to derive inductive categories of cognitive processes from reasoning traces of experts that are generating \textcolor{purple}{\{math,science\}} distractors for a multiple choice exam. Follow qualitative analysis best practices: bottom-up coding, constant comparison, memoing, and grounding all claims in the data.
\par}
\promptrule
\textbf{User Prompt} \par
\promptrule
{
\ttfamily
I will provide you with a list of reasoning traces. Your task is to discover common categories of reasoning or cognitive behaviors of the experts that are generating \textcolor{purple}{\{math,science\}} distractors, using a systematic inductive approach.

Your responsibilities:

1. Identify recurring cognitive behaviors or steps.
2. For every category, provide:
   - A clear definition (1--2 sentences)
   - A description of what behaviors fall under it
3. Provide 2--3 grounded citations for each category:
   - Verbatim excerpts from the traces
   - Include trace ID and step number (or text location)
4. Do not invent anything not present in the traces.
5. Focus only on recurring categories, not one-off behaviors.

OUTPUT FORMAT:

\#\# Discovered Categories

\#\#\# Category 1: [Name]
**Definition:**
[...]

**Example Citations:**
- "..." (Trace X, Step Y)
- "..." (Trace A, Step B)
- "..." (Trace C, Step D)

\#\#\# Category 2: [Name]
**Definition:**
[...]

(Continue as needed.)

After listing categories, include:

\#\# Notes on Method \& Coverage
Explain how the categories were derived and how representative they are.

---

Here are the reasoning traces:

\promptvar{\{traces\_concatenated\}}
\par}
\end{promptbox}
\promptcaption{Prompt used for open coding of expert reasoning traces. \promptvar{\{traces\_concatenated\}} is replaced with 10 double-new-line concatenated reasoning traces of either \ds or \glm generating distractors when prompted with the neutral prompt, as described in \cref{sec:taxonomy-method-details}. We use \textcolor{purple}{math} for Eedi and \textcolor{purple}{science} for SciQ.}{tab:prompt-open-coding}

\subsection{Annotation of Traces}
\label{app:method-annotation}
As outlined in \cref{sec:method-annotation}, we annotate all reasoning traces separately. Each trace represents an LLM's reasoning process as it generated three distractors for a single math or science problem.

\paragraph{Automatic annotation.}
We use \gptfourone~\citep{openai2024gpt4technicalreport} with greedy decoding and maximum output length of 16k tokens as the annotation model.
Our annotation pipeline (\cref{alg:annotation-pipeline}) proceeds in four steps: we first extract trace-specific grounded examples for each tag from the full trace, then split the trace into chunks, annotate each chunk (in batches) using the taxonomy and grounded examples, and finally concatenate the annotated chunks.

A single-pass annotation over the entire trace yields poor results due to length. Chunking addresses this, but individual chunks may lack problem-specific context established earlier in the trace, e.g., ``This gets us $2+9m$'' could be either \texttt{INST} or \texttt{CORR} without knowing the correct answer. To compensate, we extract grounded examples from the full trace and include them in the annotation prompt for every chunk, promoting consistent decision boundaries across chunks.

\begin{algorithm*}
\caption{Annotation pipeline applied to a single reasoning trace, i.e., one LLM generating three distractors for one MCQ.}
\label{alg:annotation-pipeline}
\begin{algorithmic}[1]
\Require trace $t$ (LLM reasoning for one MCQ's three distractors), taxonomy $\mathcal{T}$
\State $E \gets \textsc{ExtractExamples}(t, \mathcal{T})$ \Comment{up to 3 spans per tag}
\State $C \gets \textsc{ChunkText}(t)$ \Comment{split on double line breaks}
\State $A \gets [\;]$
\For{batch $B$ in $C$}
    \State $A.\textsc{extend}(\textsc{AnnotateChunks}(B, E, \mathcal{T}))$
\EndFor
\State \Return $\textsc{Concat}(A)$
\end{algorithmic}
\end{algorithm*}

\textbf{Example extraction.} We prompt the LLM with the taxonomy and the complete trace (\cref{tab:examples_prompt}), instructing it to extract up to three representative spans per tag. Tags that do not appear in the trace are omitted from the example set.

\textbf{Chunking.} We split the trace on double line breaks once a segment exceeds 500 characters, applying a hard split if no double line break occurs within 2000 characters. Double line breaks were chosen because models typically use them to separate semantic units; the 500/2000 thresholds balance segment length against annotation quality.

\textbf{Per-chunk annotation.} We annotate each chunk individually, conditioned on the grounded examples and a dataset-specific description of our taxonomy for Eedi (\cref{tab:prompt-taxonomy-eedi}) and SciQ (\cref{tab:prompt-taxonomy-sciq}), used as part of the annotation prompt (\cref{tab:annotation_prompt}). While the grounding is adapted to dataset-specific examples, the underlying decision boundaries remain consistent across descriptions.

\textbf{Stratified sampling.} We annotate $885$ traces in total. For SciQ, the higher-performance stratification bucket comprises all $165$ available traces and the lower-performance bucket comprises a random subset of $240$ traces. For Eedi, both buckets contain $240$ traces each.

\vspace{20pt}
\begin{promptbox}
\textbf{System Prompt} \par
\promptrule
{
\ttfamily
You are a helper that extracts up to 3 short example spans for each taxonomy label from a single reasoning trace.

For each tag, return a short examples block (plain text) with one line per tag in the following form:
\texttt{\textless TAG\textgreater: example1; example2}

If no examples exist for a tag, use: \texttt{\textless TAG\textgreater: (none)}

Return only the plain text block (no JSON, no commentary).

TAXONOMY:
\promptvar{\{taxonomy\_description\}}
\par}
\promptrule
\textbf{User Prompt} \par
\promptrule
{
\ttfamily
TRACE START\par
\promptvar{\{trace\}}\par
TRACE END

Return only the examples block as described.
\par}
\end{promptbox}
\promptcaption{Prompt for extracting short example spans for each strategy of the taxonomy, if present, from a reasoning trace \promptvar{\{trace\}}. \promptvar{\{taxonomy\_description\}} is replaced with the taxonomy description in \cref{tab:prompt-taxonomy-eedi} for Eedi and \cref{tab:prompt-taxonomy-sciq} for SciQ, respectively.}{tab:examples_prompt}

\begin{promptbox}
\textbf{System Prompt} \par
\promptrule
{
\ttfamily
You are annotating a chunk of a thinking-out-loud protocol produced by an expert model with markers of a taxonomy.

Context: \par
The text is a verbalized reasoning trace of an expert generating incorrect distractor answers for a \textcolor{purple}{\{math,science\}} multiple-choice question.

The expert's task was: \par
``You will be given a \textcolor{purple}{\{math,science\}} question. Please generate 3 incorrect distractor answers for the question to be used as multiple-choice options in a multiple-choice exam.''

The protocol contains the expert's internal reasoning, planning, and candidate generation steps. \par
Your job is to annotate this text by inserting taxonomy tags.

Each marker marks the END of the smallest possible span that instantiates the category.

TAXONOMY \par
\promptvar{\{taxonomy\_description\}}

EXAMPLES: \par
\promptvar{\{extracted\_examples\}}
\par}
\promptrule
\textbf{User Prompt} \par
\promptrule
{
\ttfamily
CHUNK START \par
\promptvar{\{trace\_chunk\}} \par
CHUNK END

Return only the annotated chunk (no explanations).
\par}
\end{promptbox}
\promptcaption{Prompt for annotating chunks of an LLM's reasoning when generating distractors with strategies from our taxonomy. \promptvar{\{taxonomy\_description\}} is replaced with the taxonomy description from \cref{tab:prompt-taxonomy-eedi} for Eedi and \cref{tab:prompt-taxonomy-sciq} for SciQ. \promptvar{\{extracted\_examples\}} is replaced with the examples extracted using the prompt in \cref{tab:examples_prompt}. \promptvar{\{trace\_chunk\}} is replaced with a chunk of the LLM's distractor generation reasoning trace. We use \textcolor{purple}{math} for Eedi and \textcolor{purple}{science} for SciQ.}{tab:annotation_prompt}

\begin{promptbox}
  \textbf{Taxonomy Description (Eedi)} \par
\promptrule
  {
  \ttfamily
  <INTER> \par
  Definition: Reasoning about the task instructions or requirements --- what the question asks for and what counts as valid
  answers. \par
  Rules:
  - Only mark when the expert revisits the task description and subsequently tries to gain clarity about the task itself.
  - Do NOT mark execution steps, calls to produce output, or listing candidates (e.g., "I'll produce:", "Let's do:",
  "Distractor1: 0.4\textless INST\textgreater"). \par
  Examples:
  - "We are given the question: ..."
  - "However\textless RECON\textgreater, the task is to generate three incorrect distractors, not the correct answer\textless
  INTER\textgreater" \par

  <LINK> \par
  Definition: Naming a conceptual axis or category from which candidate distractors are drawn. \par
  Rules:
  - Mark only the axis-preamble in an "axis: item1, item2, ..." enumeration. Each named item still gets \textless
  INST\textgreater.
  - Do NOT mark output-format/rendering meta (untagged): "options should be in decimal form", "options formatted similarly to
  the correct answer", "I should output exactly:", "use the template".
  - Do NOT mark error-category headers like "Common mistakes:" or "Common errors students make:" --- these introduce ERR\_DESC
  content.
  - Rare in math traces; most distractors come from error simulation (ERR\_DESC/ERR\_SIM), not conceptual-category enumeration.

  <CORR> \par
  Definition: Correct computation or reasoning toward the correct solution for the question. \par
  Rules:
  - Mark whenever correct reasoning or the correct answer is referenced.
  - If correct reasoning and errors are discussed together, mark both. \par
  Examples:
  - "2 ÷ 1/5 = 10\textless CORR\textgreater"
  - "Multiplying both sides by 4 gives 20 = k\textless CORR\textgreater, but a student might only multiply the
  numerator\textless ERR\_DESC\textgreater" \par

  <ERR\_DESC> \par
  Definition: High-level verbal description of a common mistake or misconception. \par
  Rules:
  - Mark every description of an error. \par
  Examples:
  - "A common mistake is forgetting to flip the fraction\textless ERR\_DESC\textgreater"
  - "46 \textless INST\textgreater (forgetting to add 2)\textless ERR\_DESC\textgreater"
  - "(x,y)=(-2,15)\textless INST\textgreater [from sign error\textless ERR\_DESC\textgreater]"
  - "Mis-handling the negative\textless ERR\_DESC\textgreater: -10 + 8 \textless ERR\_SIM\textgreater = 2\textless
  INST\textgreater" \par

  <ERR\_SIM> \par
  Definition: Explicitly simulating incorrect reasoning. \par
  Rules:
  - Mark when the expert simulates an incorrect calculation.
  - Single incorrect equations can be marked if they represent erroneous reasoning.
  - Mark the final incorrect outcome with \textless INST\textgreater.
  - ERR\_DESC = a high level error description; ERR\_SIM = a specific execution of an error \par
  Examples:
  - "5 - 2 = 3, then add 1 = \textless ERR\_SIM\textgreater 4\textless INST\textgreater"
  - "9 + 3 = 12, write down 2, forget to carry the 1… final result \textless ERR\_SIM\textgreater 82\textless INST\textgreater"
  - "Convert the fraction incorrectly \textless ERR\_DESC\textgreater: compute: 1 2/3 \textless ERR\_SIM\textgreater\textless
  INST\textgreater" \par

  <INST> \par
  Definition: Any incorrect outcome (number, symbol, expression). \par
  Rules:
  - Mark every candidate, even if later rejected.
  - Mark candidate values even when they appear inside task interpretation or reconsideration spans, as long as they name
  concrete answer options.
  - Each value in an enumeration of candidates is marked separately; enumeration markers like 1., 2., 3. are NOT tagged. \par
  Examples:
  - "0.4\textless INST\textgreater, 0.1\textless INST\textgreater, 2.5\textless INST\textgreater"
  - "Possible answers could be Alice \textless INST\textgreater, Bob \textless INST\textgreater, etc"
  - "980\textless INST\textgreater might work"
  - "(x,y)=(-2,15)\textless INST\textgreater [from sign error\textless ERR\_DESC\textgreater]" \par

  <PLAUS> \par
  Definition: Judgment of how likely a student would choose an error or candidate. \par
  Rules:
  - Mark plausibility comparisons or checks for incorrectness.
  - If also about final set, mark both PLAUS and CURATE. \par
  Examples:
  - "0.4\textless INST\textgreater is more plausible than 0.1\textless INST\textgreater\textless PLAUS\textgreater"
  - "The student forgets to add?\textless ERR\_DESC\textgreater Plausibly\textless PLAUS\textgreater"
  - "0.4\textless INST\textgreater is not a good distractor\textless PLAUS\textgreater"
  - "But is a student going to make that mistake?\textless PLAUS\textgreater" \par

  <DISCR> \par
  Definition: Verifying a candidate is incorrect / distinct from the correct answer (or rejecting one that's too close). \par
  Rules:
  - Mark the verification act, not the candidate itself.
  - Triggers: distinctness checks ("all different", "X is correct, so we need incorrect"), per-candidate "but it's not the
  answer" tail-clauses (mid-sentence split), "too close to correct" rejections.
  - Distinct from PLAUS (would a student pick it?) and CURATE (final-set quality).
  - Do NOT fire on general "doesn't make sense" / "not sure" dismissal or on shape-based rejection ("not an integer"). DISCR
  asserts incorrectness or insufficient distinctness from the correct answer. \par
  Examples:
  - "0.256 is correct, so we need incorrect ones.\textless DISCR\textgreater"
  - "8/3 $\approx$ 2.66 is too close to 2.828 and 2.\textless DISCR\textgreater" \par

  <CURATE> \par
  Definition: Evaluation or selection of the final set of distractors (coverage, diversity, redundancy). \par
  Rules:
  - Only mark when reasoning explicitly concerns the final set.
  - Otherwise, mark PLAUS. \par
  Examples:
  - "Keep 0.4\textless INST\textgreater and 2.5\textless INST\textgreater, drop 0.1\textless INST\textgreater to cover error
  types\textless CURATE\textgreater"
  - "0.4\textless INST\textgreater seems plausible\textless PLAUS\textgreater, keep that\textless CURATE\textgreater" \par

  <RECON> \par
  Definition: Reconsideration of a previous interpretation, candidate, plausibility judgment, or curation decision. \par
  Rules:
  - Place \textless RECON\textgreater immediately after the cue word indicating reconsideration.
  - Marks the act of reconsidering, not the outcome.
  - Common cues: "actually", "alternatively", "instead", "however", "but wait", "on second thought", "reconsider" \par
  Examples:
  - "Actually\textless RECON\textgreater, ..."
  - "Alternatively\textless RECON\textgreater, 980\textless INST\textgreater could work\textless PLAUS\textgreater"
  - "On second thought\textless RECON\textgreater, that distractor is not likely\textless PLAUS\textgreater"
\par}
\end{promptbox}
\promptcaption{The description of our taxonomy used for Eedi (\cref{tab:taxonomy-definition}) including rules and examples. We use this
description as part of both the annotation (\cref{tab:annotation_prompt}) and example extraction prompts
(\cref{tab:examples_prompt}).}{tab:prompt-taxonomy-eedi}

\begin{promptbox}
\textbf{Taxonomy Description (SciQ)} \par
\promptrule
{
\ttfamily
\# Cross-cutting rules (abbreviated) \par
CC1. Tag every occurrence (no first-mention-only). CC2. Stack tags only when each genuinely applies. CC3. Output-formatting / template / phrasing talk is untagged [\ldots]. CC4. Same fact gets different tags by function (e.g.\ \textless CORR\textgreater\ when grounding the answer vs.\ \textless LINK\textgreater\ when defining the design space). CC5. Place tags at clause boundaries, not at sentence ends. CC6. Pivot cues at sentence start (\textit{Instead,} \textit{Alternatively,} \textit{Actually,} \textit{Wait,} \textit{Another idea:}, \ldots) ALWAYS trigger \textless RECON\textgreater\ immediately after the cue. \par

\# Tags \par

\textless INTER\textgreater\ --- Engaging with what the task/question asks (interpreting, restating, breaking down). Switch to \textless CORR\textgreater\ once solving begins; section headers and output-formatting are not INTER. \par
Ex: "The question asks for the primary product of photosynthesis.\textless INTER\textgreater" \par

\textless CORR\textgreater\ --- States, deduces, or grounds the correct answer. Place on the specific term span when named inline. Hedging meta toward committing to the answer is CORR. \par
Ex: "Glucose is the primary product of photosynthesis.\textless CORR\textgreater" \par

\textless LINK\textgreater\ --- Conceptual/categorical design space for distractors, with NO specific candidate named. Specificity test: replace the noun with "such things"; if still a category-level criterion, it is LINK. Brainstorm-intro phrases ("Possible distractors could include:") are untagged, not LINK. \par
Ex: "Distractors should be other types of bonds.\textless LINK\textgreater" \par

\textless INST\textgreater\ --- A specific value/term/concept named as a concrete answer-shaped option (drafted, brainstormed, or in a taxonomic enumeration). Each item in an enumeration is tagged separately. Axis-then-items bullets: axis gets \textless LINK\textgreater, each item gets \textless INST\textgreater. A correct-answer term in a candidate list still gets \textless INST\textgreater\ (plus \textless CORR\textgreater). [\ldots] \par
Ex: "- By causative agent\textless LINK\textgreater: bacterial\textless INST\textgreater, viral\textless INST\textgreater, parasitic\textless INST\textgreater, fungal\textless INST\textgreater." \par

\textless PLAUS\textgreater\ --- Why a \textit{specific candidate} would be tempting (similarity, familiarity, commonness). Requires a named candidate (general "distractors should be plausible" is LINK). Common-misconception spans stack \textless ERR\_DESC\textgreater\textless PLAUS\textgreater. Dismissive scoping ("but not in the leg") is untagged. \par
Ex: "mitosis\textless INST\textgreater\ (a common confusion\textless PLAUS\textgreater)" \par

\textless DISCR\textgreater\ --- Why a \textit{specific candidate} is incorrect or contrasts with the correct answer. General "distractors must be wrong" is LINK; whole-set sanity checks are CURATE. Same dismissive-scoping caveat as PLAUS. \par
Ex: "It sounds plausible because of x\textless PLAUS\textgreater\ but is wrong due to y.\textless DISCR\textgreater" \par

\textless ERR\_DESC\textgreater\ --- High-level description of a cognitive error/misconception. The span MUST contain an action verb (confuse, mix up, forget, misapply, associate, \ldots); pure factual statements don't qualify. \par
Ex: "A student might associate mitosis with cell growth and forget it produces two identical daughter cells.\textless ERR\_DESC\textgreater" \par

\textless ERR\_SIM\textgreater\ --- Step-by-step simulation of a misconception (vs.\ ERR\_DESC's high-level description). \par
Ex: "A student might think: photosynthesis needs light, mitochondria use light too\textless ERR\_SIM\textgreater, so the answer is mitochondria\textless INST\textgreater." \par

\textless CURATE\textgreater\ --- Selecting/refining/finalizing the distractor set, including informal commits, end-of-trace selection, set-level sanity checks, and the closing line of a bare final-list block. Brainstorm-introduction phrases are NOT CURATE. \par
Ex: "Let's pick the three strongest ones.\textless CURATE\textgreater" \par

\textless RECON\textgreater\ --- Reconsidering a prior interpretation/candidate/judgment/decision. Placed immediately after the pivot cue (never trailing). New content after \textless RECON\textgreater\ is tagged on its own. \par
Ex: "Instead\textless RECON\textgreater, I'll try a different category.\textless LINK\textgreater" \par

\# Decision tree (top-down; first match wins; co-tag only when each genuinely applies) \par
1. Reconsideration cue at sentence start? $\rightarrow$ \textless RECON\textgreater, then continue. \quad
2. Names a specific candidate? $\rightarrow$ \textless INST\textgreater, then continue. \quad
3. Why a candidate is tempting? $\rightarrow$ \textless PLAUS\textgreater. \quad
4. Why a candidate is incorrect? $\rightarrow$ \textless DISCR\textgreater. \quad
5. Cognitive error with action verb? $\rightarrow$ \textless ERR\_DESC\textgreater\ (procedural $\rightarrow$ \textless ERR\_SIM\textgreater). \quad
6. Selecting/refining the final set? $\rightarrow$ \textless CURATE\textgreater. \quad
7. Design-space, no candidate named? $\rightarrow$ \textless LINK\textgreater. \quad
8. Engaging with the task? $\rightarrow$ \textless INTER\textgreater. \quad
9. Grounding the correct answer? $\rightarrow$ \textless CORR\textgreater. \quad
Otherwise $\rightarrow$ untagged (output-formatting always falls here, per CC3).
\par}
\end{promptbox}
\promptcaption{The SciQ taxonomy prompt, abbreviated. Long enumerations of trigger phrases, sub-patterns, and operational examples are elided as [\ldots]; the full prompt is released with the code. We use this
description as part of both the annotation (\cref{tab:annotation_prompt}) and example extraction prompts
(\cref{tab:examples_prompt}).}{tab:prompt-taxonomy-sciq}

\paragraph{Annotation quality.}
We report detailed annotation performance---comparing agreement between human and \gptfourone's annotations---per strategy, on distractor-generation-traces of \ds and \glm, across both datasets, and CoT versus reasoning mode in \cref{tab:annotation_evaluation}.
Manual analyses were conducted on 64 traces, uniformly stratified across dataset, model, prompt, and high- versus low-performance cases. As a result, strategies that the models rarely engage in have very low support. However, the resulting uncertainty does not meaningfully affect the analyses in \cref{sec:experiments-process-analysis}, since such strategies are far from playing a dominant role in the frequency analysis or dominant paths in the subprocess analysis.
Overall agreement was high across most strategies.

\begin{table*}[!htbp]
\centering
\renewcommand{\arraystretch}{1.15}
\small
\begin{tabular}{lll|ccc|ccc}
\toprule
 &  &  & \multicolumn{3}{c|}{\textbf{Eedi}} & \multicolumn{3}{c}{\textbf{SciQ}} \\
\cline{4-9}
\textbf{Model} & \textbf{Prompt} & \textbf{Strategy} & \textbf{\#} & \textbf{Precision} & \textbf{Recall} & \textbf{\#} & \textbf{Precision} & \textbf{Recall} \\
\midrule
\multirow{20}{*}{\textbf{\dsabbr}}
& \multirow{10}{*}{\textbf{CoT}} & Task Interpretation & 13 & 1.00 & 1.00 & 24 & 1.00 & 1.00 \\
& & Correct Answer Ref. & 13 & 1.00 & 1.00 & 23 & 1.00 & 1.00 \\
& & Conceptual Link & 0 & -- & -- & 15 & 0.93 & 0.93 \\
& & Error Description & 48 & 0.98 & 0.89 & 11 & 0.64 & 0.78 \\
& & Error Simulation & 9 & 1.00 & 1.00 & 0 & -- & -- \\
& & Outcome Instantiation & 50 & 1.00 & 0.94 & 66 & 1.00 & 0.97 \\
& & Plausibility Check & 4 & 0.75 & 1.00 & 29 & 1.00 & 0.97 \\
& & Discriminability Check & 2 & 1.00 & 1.00 & 30 & 0.97 & 0.97 \\
& & Final Set Curation & 5 & 1.00 & 0.71 & 7 & 1.00 & 0.88 \\
& & Reconsideration & 1 & 1.00 & 1.00 & 1 & 1.00 & 0.50 \\
\cline{2-9}
& \multirow{10}{*}{\textbf{Reasoning}} & Task Interpretation & 63 & 0.87 & 1.00 & 43 & 0.91 & 1.00 \\
& & Correct Answer Ref. & 88 & 1.00 & 0.96 & 54 & 1.00 & 1.00 \\
& & Conceptual Link & 0 & -- & -- & 29 & 0.72 & 1.00 \\
& & Error Description & 273 & 1.00 & 0.99 & 6 & 1.00 & 0.86 \\
& & Error Simulation & 71 & 0.99 & 0.97 & 0 & -- & -- \\
& & Outcome Instantiation & 352 & 1.00 & 0.99 & 176 & 1.00 & 1.00 \\
& & Plausibility Check & 94 & 1.00 & 0.98 & 38 & 1.00 & 0.95 \\
& & Discriminability Check & 43 & 0.88 & 1.00 & 47 & 0.96 & 0.83 \\
& & Final Set Curation & 28 & 1.00 & 1.00 & 37 & 1.00 & 0.97 \\
& & Reconsideration & 46 & 1.00 & 0.87 & 4 & 1.00 & 0.80 \\
\midrule
\multirow{20}{*}{\textbf{\glm}}
& \multirow{10}{*}{\textbf{CoT}} & Task Interpretation & 7 & 1.00 & 1.00 & 12 & 0.83 & 1.00 \\
& & Correct Answer Ref. & 17 & 1.00 & 0.94 & 14 & 1.00 & 1.00 \\
& & Conceptual Link & 0 & -- & -- & 2 & 1.00 & 0.50 \\
& & Error Description & 103 & 1.00 & 0.98 & 8 & 1.00 & 1.00 \\
& & Error Simulation & 28 & 1.00 & 1.00 & 0 & -- & -- \\
& & Outcome Instantiation & 88 & 1.00 & 0.98 & 47 & 1.00 & 0.85 \\
& & Plausibility Check & 12 & 1.00 & 1.00 & 33 & 0.94 & 0.97 \\
& & Discriminability Check & 5 & 1.00 & 1.00 & 25 & 0.96 & 0.89 \\
& & Final Set Curation & 3 & 1.00 & 0.75 & 6 & 1.00 & 0.75 \\
& & Reconsideration & 14 & 1.00 & 0.88 & 0 & -- & -- \\
\cline{2-9}
& \multirow{10}{*}{\textbf{Reasoning}} & Task Interpretation & 19 & 0.84 & 1.00 & 32 & 1.00 & 1.00 \\
& & Correct Answer Ref. & 27 & 1.00 & 1.00 & 18 & 1.00 & 1.00 \\
& & Conceptual Link & 0 & -- & -- & 19 & 0.95 & 1.00 \\
& & Error Description & 164 & 1.00 & 0.98 & 12 & 0.83 & 0.77 \\
& & Error Simulation & 49 & 1.00 & 1.00 & 0 & -- & -- \\
& & Outcome Instantiation & 241 & 1.00 & 0.99 & 279 & 1.00 & 1.00 \\
& & Plausibility Check & 29 & 1.00 & 0.85 & 72 & 1.00 & 0.92 \\
& & Discriminability Check & 6 & 0.83 & 1.00 & 75 & 1.00 & 0.87 \\
& & Final Set Curation & 16 & 1.00 & 0.94 & 59 & 0.95 & 0.93 \\
& & Reconsideration & 17 & 1.00 & 1.00 & 14 & 1.00 & 1.00 \\
\bottomrule
\end{tabular}
\caption{Performance of LLM-based annotations per taxonomy strategy (\cref{tab:taxonomy-definition}) for \ds (\dsabbr) and \glm, with CoT vs. reasoning, for both datasets. Note that RECON and CURATE have very little support in the CoT setting.}
\label{tab:annotation_evaluation}
\end{table*}

\subsection{Qualitative Examples}
\label{app:qualitative-examples}
We give two complementary sets of examples, both drawn from \ds and \glm reasoning traces. \Cref{tab:strategy-examples} illustrates each taxonomy strategy (from \cref{tab:taxonomy-definition}) with short verbatim excerpts from the collected traces, to make the annotation concrete. \Cref{tab:distractor-examples} instead focuses on the generated \emph{distractors}, classified by whether our equivalence judge (the same one underlying proportional match, \cref{app:measure-performance}) deems them equivalent to a human-authored distractor. It contrasts these high-quality distractors with low-quality ones that match no human-authored distractor (e.g., a candidate equal to the correct answer, or one not grounded in a coherent misconception), showing the full ground-truth and model distractor sets alongside the trace excerpt that produced or justified the focal candidate. Excerpts are shortened for space, with deletions marked \texttt{[...]}. Examples are from Eedi, except the \texttt{LINK} row in \cref{tab:strategy-examples}, which is near-absent on Eedi (\cref{tab:component_presence}) and is therefore drawn from SciQ.

\begin{table*}[!htbp]
\centering
\small
\renewcommand{\arraystretch}{1.25}
\begin{tabular}{@{}p{1.5cm}p{6.4cm}p{6.4cm}@{}}
\toprule
\textbf{Strategy} & \textbf{Example 1} & \textbf{Example 2} \\
\midrule
\texttt{INTER} &
``The question asks who is correct regarding the method to find 40\% of an amount.'' &
``The question likely means: among these two numbers, which one(s) are rational?'' \\
\texttt{CORR} &
``The correct answer is $\frac{9}{20}$.'' &
``Remember, $0.5841$ rounded to one significant figure [...] gives $0.6$.'' \\
\texttt{LINK} &
``[...] the options should be animal names.'' &
``These should be related to bodily movements, digestion, or similar terms [...]'' \\
\texttt{ERR\_DESC} &
``A student might think $5+x$ is the same as $5x$.'' &
``Maybe they simplify $\frac{6}{27}$ to $\frac{1}{3}$?'' \\
\texttt{ERR\_SIM} &
``If she adds 1 to both sides (incorrectly): $\frac{5x}{3} = 3$.'' &
``But if you think addition comes before subtraction, you might do $8+5=13$, then $9-13=-4$.'' \\
\texttt{INST} &
``Distractor3: Both Tom and Katie are correct'' &
``That gives $78{,}125$'' \\
\texttt{PLAUS} &
``These are plausible mistakes where a student might [...]'' &
``This is a plausible distractor if [...]'' \\
\texttt{DISCR} &
```Cannot be simplified' is an option, but it is incorrect because it can be simplified.'' &
``$3.7150$ is essentially $3.715$. So maybe not.'' \\
\texttt{CURATE} &
``[...] This is the most robust set.'' &
``[...] I'll go with these three.'' \\
\texttt{RECON} &
``Wait, is $40$ too obvious?'' &
``But wait: Could there be other possibilities?'' \\
\bottomrule
\end{tabular}
\caption{Two representative excerpts illustrating each taxonomy strategy (\cref{tab:taxonomy-definition}), taken verbatim (up to deletions marked \texttt{[...]}) from \ds\ and \glm\ reasoning traces. All rows are from Eedi except \texttt{LINK}, illustrated on SciQ where it is prevalent.}
\label{tab:strategy-examples}
\end{table*}

\begin{table*}[!htbp]
\centering
\small
\renewcommand{\arraystretch}{1.3}
\begin{tabular}{@{}p{3.2cm}p{4.3cm}p{6.4cm}@{}}
\toprule
\textbf{Problem \newline(correct answer)} & \textbf{Distractors: gold vs.\ model} & \textbf{Trace excerpt / failure mode} \\
\midrule
\multicolumn{3}{@{}l}{\textit{High-quality (proportional match $>0.5$): at least two of three distractors match}} \\
Simplify $\frac{6+9m}{3}$ \newline (correct: $2+3m$) &
\textbf{Gold:} $6+3m$, $2+9m$, None of the above \newline \textbf{Model:} $2+9m$, $6+3m$, $5m$ &
``[...] $3(2+9m)/3 = 2+9m$'' \\
Median of $3,5,6,18,18$ \newline (correct: $6$) &
\textbf{Gold:} $10$, $12$, $18$ \newline \textbf{Model:} $10$, $18$, $5.5$ &
``Confusing median with mean [...] The mean is $(3+5+6+18+18)/5 = 50/5 = 10$'' \\
$30$\,cl $=\,\star$\,litres \newline (correct: $0.3$) &
\textbf{Gold:} $3000$, $3$, $0.03$ \newline \textbf{Model:} $0.03$, $3$, $30$ &
``[...] some might think [centi] means $1/1000$ (like milli) [...]'' \\
\midrule
\multicolumn{3}{@{}l}{\textit{Low-quality (proportional match $<0.5$): at most one of three distractors matches}} \\
$2\sqrt{5}+2\sqrt{45}=d\sqrt{5}$ \newline (correct: $d=8$) &
\textbf{Gold:} $7$, $20$, $4$ \newline \textbf{Model:} $5$, $6$, $12$ &
\textbf{Solution recovery:} the model never reaches $d=8$, so its values are ungrounded. \\
$232 \div 5$, remainder as a decimal \newline (correct: $0.4$) &
\textbf{Gold:} $0.2$, $2$, $0.25$ \newline \textbf{Model:} $46$, $4$, $0.4$ &
\textbf{Discriminability:} $0.4$ is the correct answer, offered as a distractor. \\
\bottomrule
\end{tabular}
\caption{High- vs.\ low-quality generated distractors from \ds/\glm\ reasoning traces on Eedi, grouped by proportional match ($>0.5$ vs.\ $<0.5$, the split used throughout the paper). Each row stacks the human-authored gold distractors over the model's three proposed distractors, and gives either the verbatim trace excerpt that produced the matching candidate (high-quality; up to deletions \texttt{[...]}) or the failure mode it exhibits (low-quality; \cref{sec:discussion}).}
\label{tab:distractor-examples}
\end{table*}

\section{Full Results}

\label{app:full-results}
Here, we present the full results for \glm and for the CoT prompting setting in the experiments from \cref{sec:experiments-performance} and \cref{sec:experiments-process-analysis}. We also briefly compare these results to those of \ds in reasoning mode.

\subsection{Performance}
\label{app:full-results-performance}
In \cref{app:distr-performance-full} we observe that both reasoning and CoT tend to reduce \#correct and \#repetitions over direct generation. On SciQ, \#repetitions remain consistently low across settings.

\begin{table*}
\centering
\renewcommand{\arraystretch}{1.15}
\small
\begin{tabular}{ll|cc|cc}
\toprule
\multirow{2}{*}{\textbf{Model}} & \multirow{2}{*}{\textbf{Prompt}}
& \multicolumn{2}{c}{\textbf{Eedi}}
& \multicolumn{2}{c}{\textbf{SciQ}} \\
\cline{3-6}
& & \textbf{\#correct} & \textbf{\#repetitions}
  & \textbf{\#correct} & \textbf{\#repetitions} \\
\midrule

\multirow{3}{*}{\textbf{\ds}}
& Direct     & 0.42 $\pm$ 0.05 & 0.19 $\pm$ 0.06 & 0.06 $\pm$ 0.02 & 0.00 $\pm$ 0.00 \\
& CoT        & 0.07 $\pm$ 0.02 & 0.02 $\pm$ 0.01 & 0.03 $\pm$ 0.02 & 0.01 $\pm$ 0.01 \\
& Reasoning  & 0.04 $\pm$ 0.02 & 0.02 $\pm$ 0.01 & 0.02 $\pm$ 0.01 & 0.02 $\pm$ 0.01 \\
\midrule

\multirow{3}{*}{\textbf{\glm}}
& Direct     & 0.57 $\pm$ 0.05 & 0.02 $\pm$ 0.01 & 0.15 $\pm$ 0.03 & 0.01 $\pm$ 0.01 \\
& CoT        & 0.06 $\pm$ 0.02 & 0.03 $\pm$ 0.02 & 0.02 $\pm$ 0.01 & 0.02 $\pm$ 0.01 \\
& Reasoning  & 0.04 $\pm$ 0.02 & 0.02 $\pm$ 0.02 & 0.02 $\pm$ 0.01 & 0.02 $\pm$ 0.01 \\
\bottomrule

\end{tabular}
\caption{Number of correct and repetitions (mean ± 95\% CI based on t-distribution) for \ds and \glm, on the Eedi and SciQ datasets, under different prompting conditions.}
\label{app:distr-performance-full}
\end{table*}

\subsection{Occurrences of Strategies}
\label{app:full-results-occurrences-of-strategies}
\cref{tab:component_presence_full} shows the occurrences of strategies for both models on both datasets. Comparing \glm to \ds (\cref{tab:component_presence}), we observe that:
(i) across models and datasets, the more concise CoT traces contain fewer strategy tags than the reasoning traces; and
(ii) the average ratios between CoT and reasoning are of similar magnitude for both models, except for \texttt{RECON}, \texttt{ERR\_SIM}, and \texttt{PLAUS}, where the ratio between CoT and reasoning is significantly larger for \ds than for \glm, and \texttt{CURATE}, for which the opposite trend holds.

\begin{table*}[!htbp]
\centering
\renewcommand{\arraystretch}{1.15}
\small
\begin{tabular}{ll|
                r@{.}l@{\,$\pm$\,}r@{.}l
                r@{.}l@{\,$\pm$\,}r@{.}l
                r@{.}l@{\,$\pm$\,}r@{.}l
                r@{.}l@{\,$\pm$\,}r@{.}l}
\toprule
& &
\multicolumn{8}{c}{\textbf{CoT}}
& \multicolumn{8}{c}{\textbf{Reasoning}} \\
\cmidrule(lr){3-10}
\cmidrule(lr){11-18}

\textbf{Model} & \textbf{Strategy} &
\multicolumn{4}{c}{\textbf{Eedi}}
& \multicolumn{4}{c}{\textbf{SciQ}}
& \multicolumn{4}{c}{\textbf{Eedi}}
& \multicolumn{4}{c}{\textbf{SciQ}} \\
\midrule
\multirow{10}{*}{\textbf{\ds}}

& Task Interpretation
& 2 & 08 & 0 & 33
& 2 & 45 & 0 & 26
& 11 & 22 & 1 & 38
& 5 & 84 & 0 & 80 \\

& Correct Answer Ref.
& 3 & 27 & 0 & 87
& 1 & 87 & 0 & 22
& 9 & 33 & 1 & 40
& 5 & 52 & 0 & 89 \\

& Conceptual Link
& 0 & 01 & 0 & 02
& 2 & 21 & 0 & 34
& 0 & 04 & 0 & 05
& 5 & 36 & 0 & 87 \\

& Error Description
& 6 & 93 & 1 & 08
& 0 & 56 & 0 & 21
& 22 & 83 & 2 & 70
& 0 & 78 & 0 & 31 \\

& Error Simulation
& 2 & 09 & 0 & 62
& 0 & 00 & 0 & 00
& 5 & 70 & 1 & 18
& 0 & 00 & 0 & 00 \\

& Outcome Instantiation
& 8 & 03 & 1 & 32
& 8 & 10 & 1 & 17
& 34 & 84 & 4 & 05
& 24 & 71 & 2 & 95 \\

& Plausibility Check
& 1 & 26 & 0 & 31
& 2 & 72 & 0 & 41
& 7 & 94 & 1 & 15
& 5 & 33 & 0 & 90 \\

& Discriminability Check
& 0 & 51 & 0 & 25
& 3 & 18 & 0 & 53
& 2 & 54 & 0 & 59
& 10 & 34 & 1 & 50 \\

& Final Set Curation
& 0 & 85 & 0 & 17
& 1 & 42 & 0 & 26
& 3 & 34 & 0 & 39
& 5 & 83 & 0 & 60 \\

& Reconsideration
& 0 & 67 & 0 & 22
& 0 & 16 & 0 & 11
& 5 & 46 & 0 & 71
& 1 & 05 & 0 & 23 \\

\midrule

\multirow{10}{*}{\textbf{\glm}}

& Task Interpretation
& 1 & 16 & 0 & 18
& 1 & 89 & 0 & 15
& 5 & 31 & 1 & 31
& 4 & 18 & 0 & 34 \\

& Correct Answer Ref.
& 4 & 99 & 3 & 45
& 1 & 67 & 0 & 17
& 9 & 39 & 3 & 09
& 3 & 72 & 0 & 55 \\

& Conceptual Link
& 0 & 01 & 0 & 01
& 0 & 79 & 0 & 27
& 0 & 03 & 0 & 04
& 1 & 86 & 0 & 43 \\

& Error Description
& 24 & 68 & 32 & 08
& 1 & 22 & 0 & 32
& 27 & 45 & 9 & 20
& 2 & 34 & 0 & 44 \\

& Error Simulation
& 14 & 39 & 20 & 74
& 0 & 00 & 0 & 00
& 14 & 41 & 7 & 86
& 0 & 00 & 0 & 00 \\

& Outcome Instantiation
& 8 & 09 & 1 & 80
& 8 & 40 & 1 & 52
& 34 & 97 & 10 & 28
& 39 & 54 & 4 & 64 \\

& Plausibility Check
& 1 & 18 & 0 & 37
& 4 & 65 & 0 & 50
& 4 & 58 & 0 & 91
& 10 & 32 & 1 & 42 \\

& Discriminability Check
& 0 & 21 & 0 & 13
& 4 & 67 & 0 & 69
& 0 & 96 & 0 & 32
& 10 & 20 & 1 & 60 \\

& Final Set Curation
& 0 & 56 & 0 & 21
& 1 & 30 & 0 & 30
& 3 & 96 & 0 & 63
& 7 & 20 & 1 & 14 \\

& Reconsideration
& 0 & 94 & 0 & 34
& 0 & 10 & 0 & 10
& 2 & 76 & 0 & 70
& 1 & 89 & 0 & 43 \\

\bottomrule
\end{tabular}
\caption{Average occurrences of taxonomy strategies (\cref{tab:taxonomy-definition}) in CoT and reasoning traces for each model on both datasets (mean $\pm$ 95\% CI based on t-distribution).}
\label{tab:component_presence_full}
\end{table*}

\subsection{Strategies Over Time}
\label{app:full-results-strategies-over-time}
\begin{figure*}
\centering
\vspace{-10pt}
\includegraphics[width=0.98\linewidth]{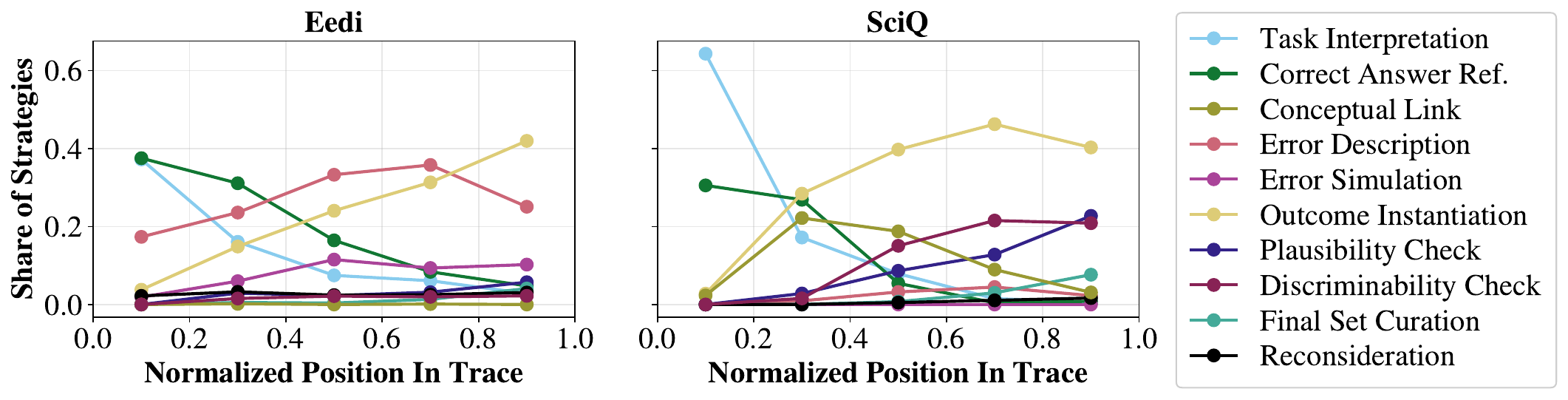}
\vspace{-5pt}
\caption{
Indicates how often each strategy in our taxonomy \cref{tab:taxonomy-definition} was annotated at different stages of \ds's CoT trace. Time is normalized (0 = start of trace, 1 = end). Note that the proportions sum to one for each of the five temporal bins. \looseness=-1
}

\label{fig:deepseek-cot-components-over-time}
\end{figure*}

\begin{figure*}
\centering

\includegraphics[width=0.98\linewidth]{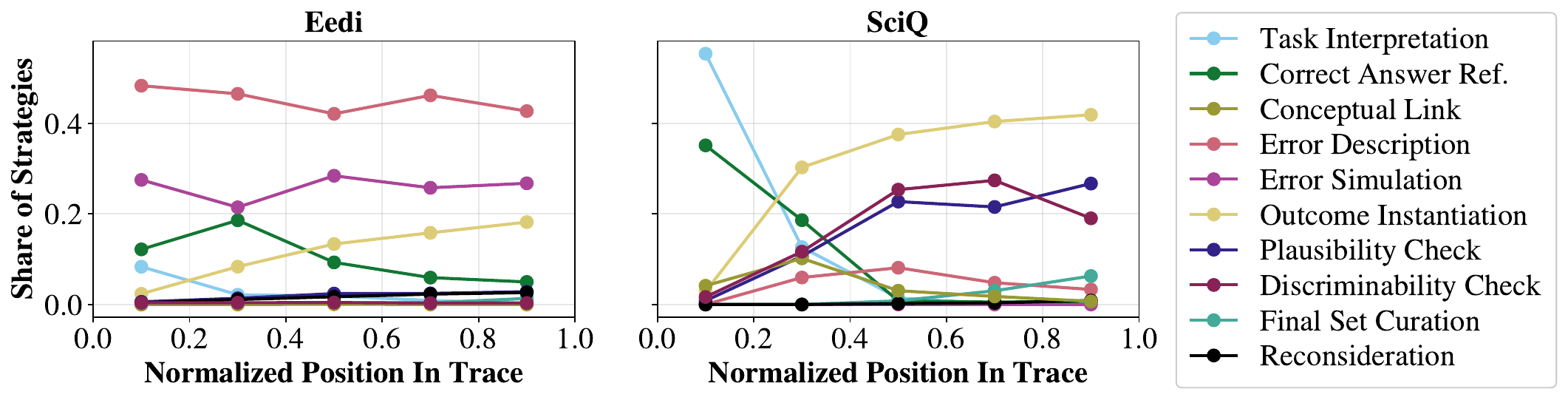}
\vspace{-5pt}
\caption{
Indicates how often each strategy in our taxonomy \cref{tab:taxonomy-definition} was annotated at different stages of \glm's CoT trace. Time is normalized (0 = start of trace, 1 = end). Note that the proportions sum to one for each of the five temporal bins. \looseness=-1
}

\label{fig:glm-cot-components-over-time}
\end{figure*}

\begin{figure*}
\centering

\includegraphics[width=0.98\linewidth]{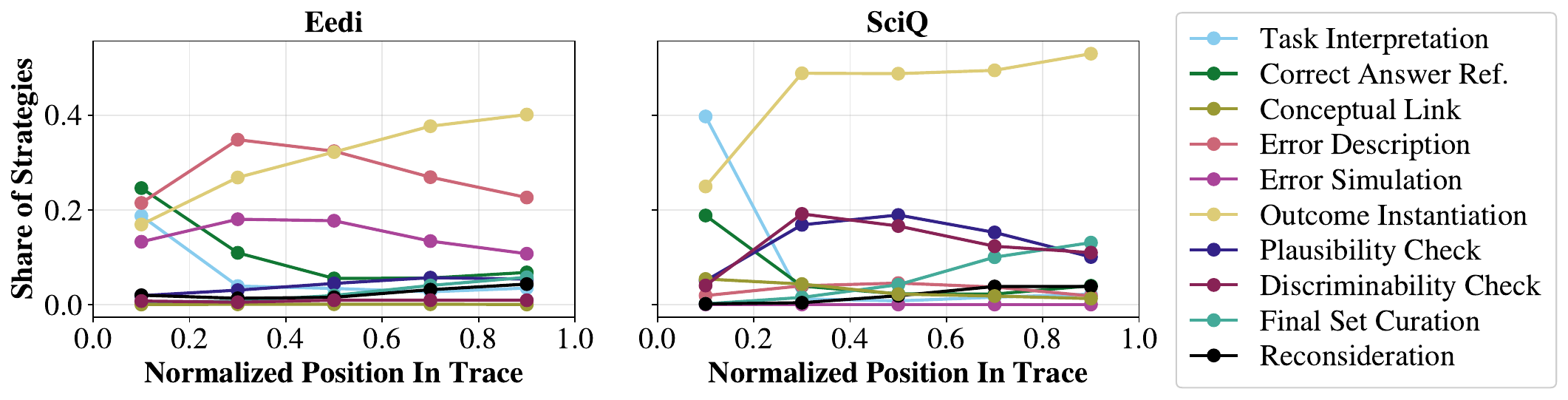}
\vspace{-5pt}
\caption{
Indicates how often each strategy in our taxonomy \cref{tab:taxonomy-definition} was annotated at different stages of \glm's reasoning trace. Time is normalized (0 = start of trace, 1 = end). Note that the proportions sum to one for each of the five temporal bins. \looseness=-1
}

\label{fig:glm-reasoning-components-over-time}
\end{figure*}

Results are shown in~\cref{fig:deepseek-cot-components-over-time} for \ds in the CoT setting, and in~\cref{fig:glm-cot-components-over-time} and~\cref{fig:glm-reasoning-components-over-time} for \glm in the CoT and reasoning settings respectively. Overall trends match those of \ds in the reasoning setting (\cref{fig:components-over-time}).
One notable difference is that \glm in CoT consistently engages in \texttt{ERR\_DESC} and \texttt{ERR\_SIM}, whereas for \ds the relative share of these strategies increases over time while \texttt{CORR} and \texttt{INTER} decrease.

\subsection{Common Sequences of Strategies}
\label{app:full-results-common-sequences-of-strategies}
\begin{figure*}[!t]
  \centering
  \begin{minipage}{0.35\textwidth}
        \includegraphics[width=\textwidth]{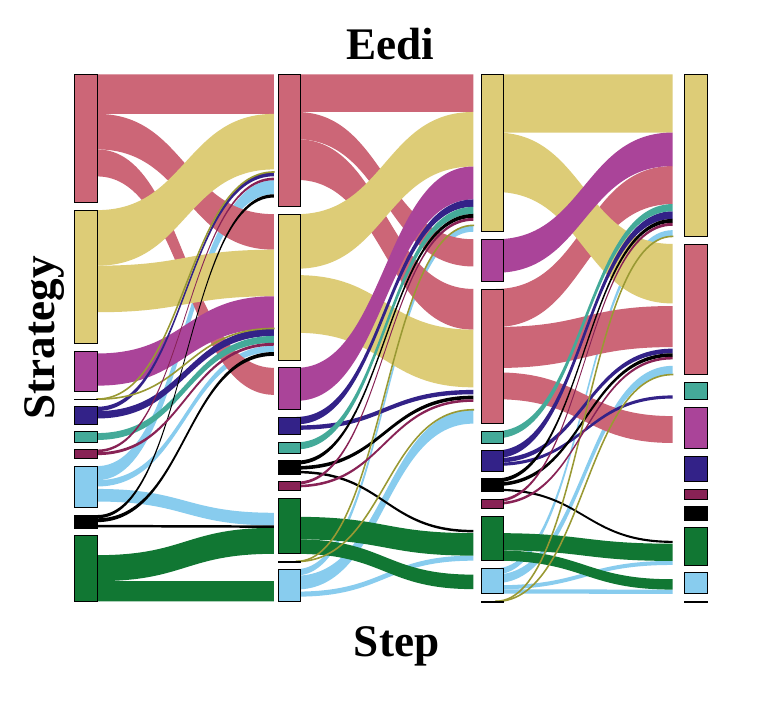}
    \end{minipage}
    \begin{minipage}{0.55\textwidth}
        \includegraphics[width=\textwidth]{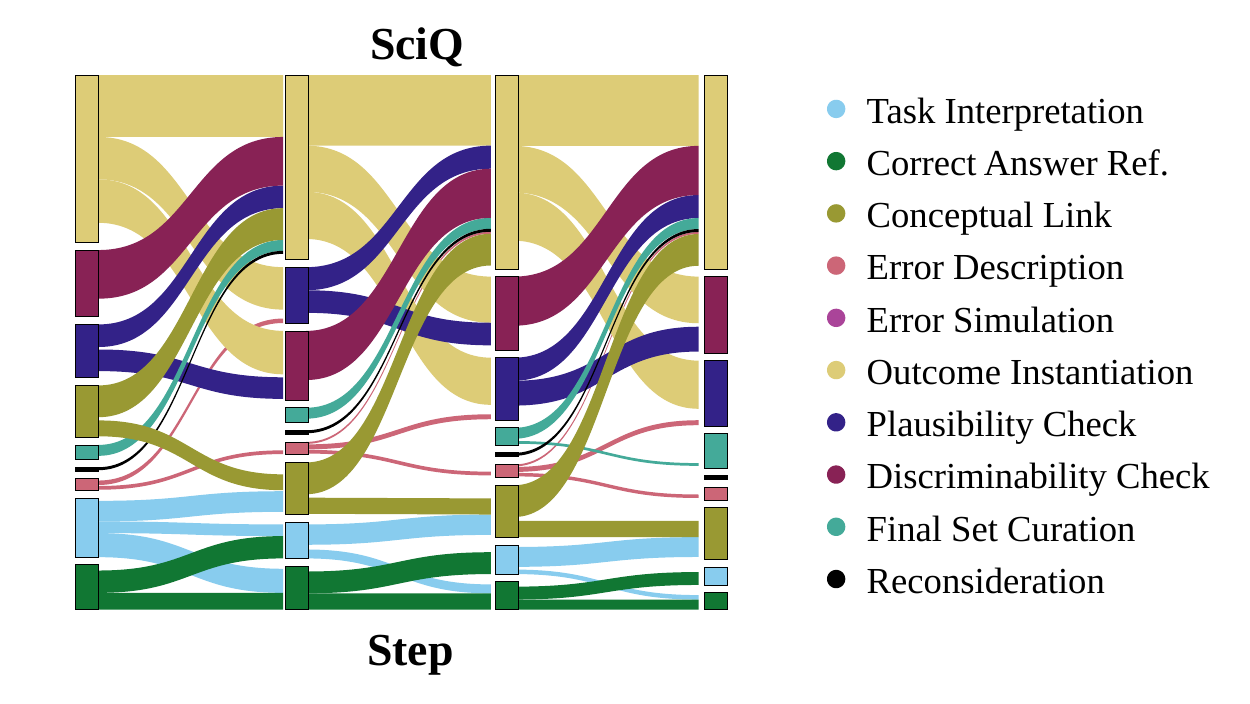}
    \end{minipage}
  \vspace{-7pt}
  \caption{
  Shows transition probabilities between strategies in \ds's CoT traces. Sequences of strategies up to length 4 (left to right). Node height represents strategy share and link widths indicate the transition probabilities between successive strategies. Only dominant (>15\% outgoing mass) transitions are visualized for simplicity.\looseness=-1
  }
  \label{fig:deepseek-cot-transition-probabilities}
  \vspace{-10pt}
\end{figure*}

\begin{figure*}[!t]
  \centering
  \begin{minipage}{0.35\textwidth}
        \includegraphics[width=\textwidth]{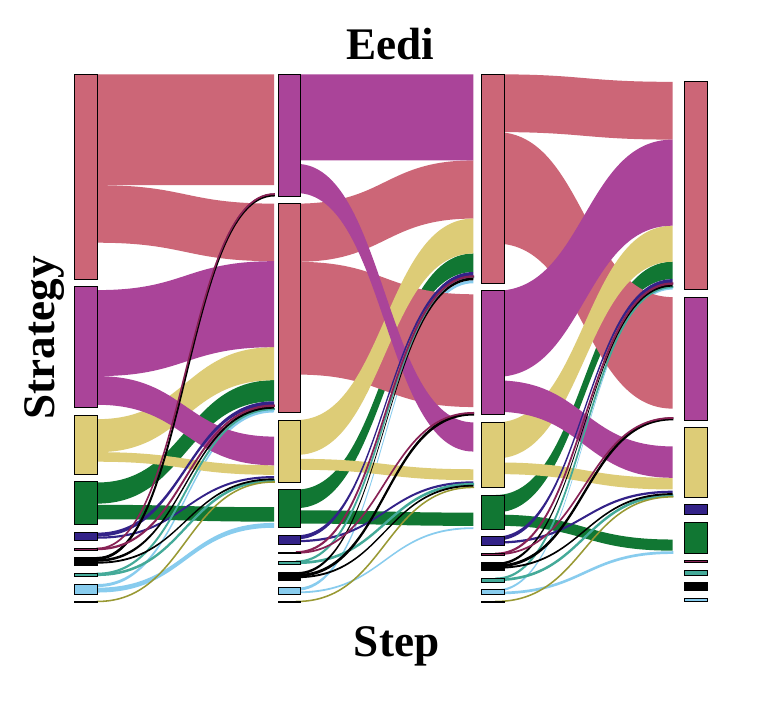}
    \end{minipage}
    \begin{minipage}{0.55\textwidth}
        \includegraphics[width=\textwidth]{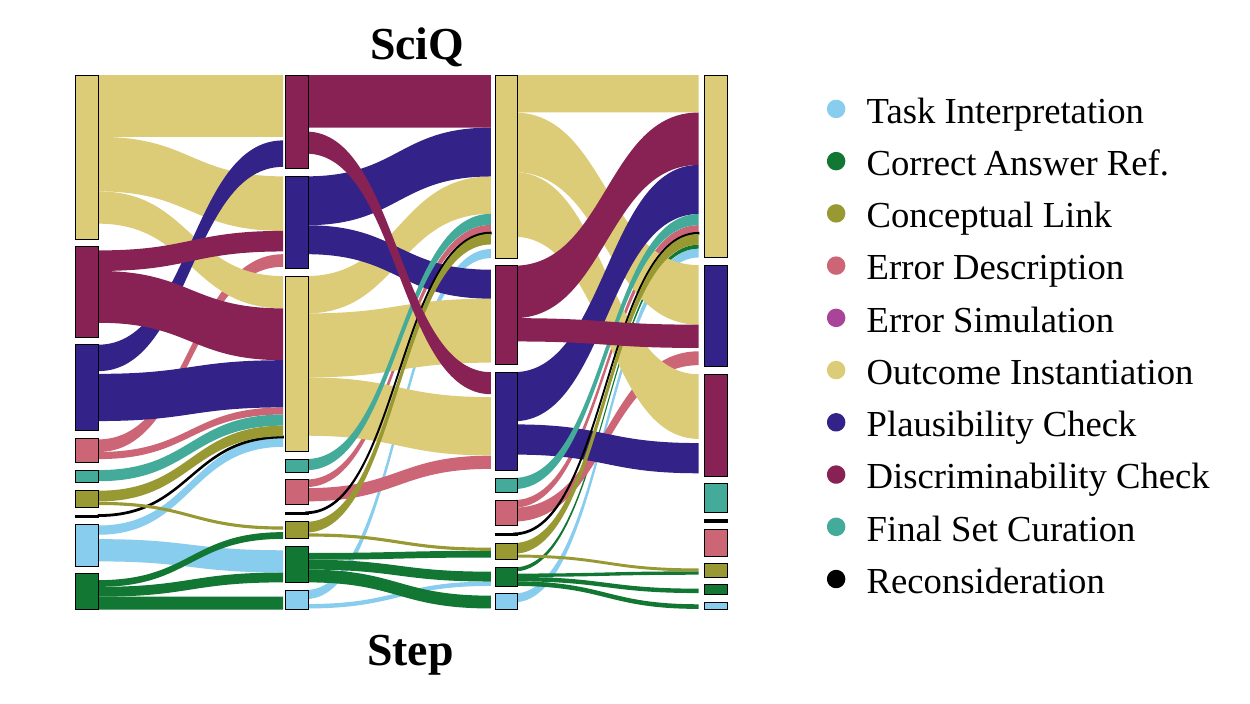}
    \end{minipage}
  \vspace{-7pt}
  \caption{
  Shows transition probabilities between strategies in \glm's CoT traces. Sequences of strategies up to length 4 (left to right). Node height represents strategy share and link widths indicate the transition probabilities between successive strategies. Only dominant (>15\% outgoing mass) transitions are visualized for simplicity.\looseness=-1
  }
  \label{fig:glm-cot-transition-probabilities}
  \vspace{-10pt}
\end{figure*}

\begin{figure*}[!t]
  \centering
  \begin{minipage}{0.35\textwidth}
        \includegraphics[width=\textwidth]{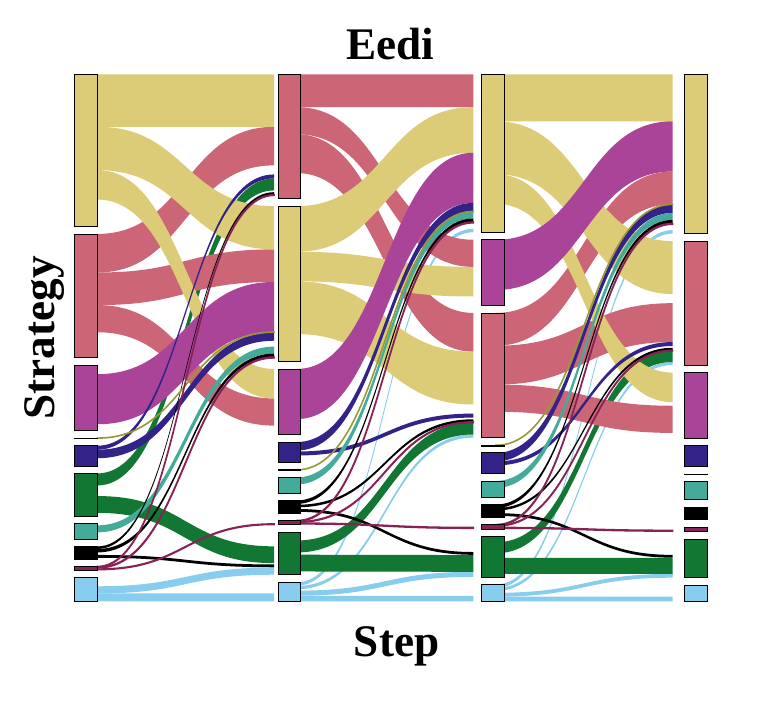}
    \end{minipage}
    \begin{minipage}{0.55\textwidth}
        \includegraphics[width=\textwidth]{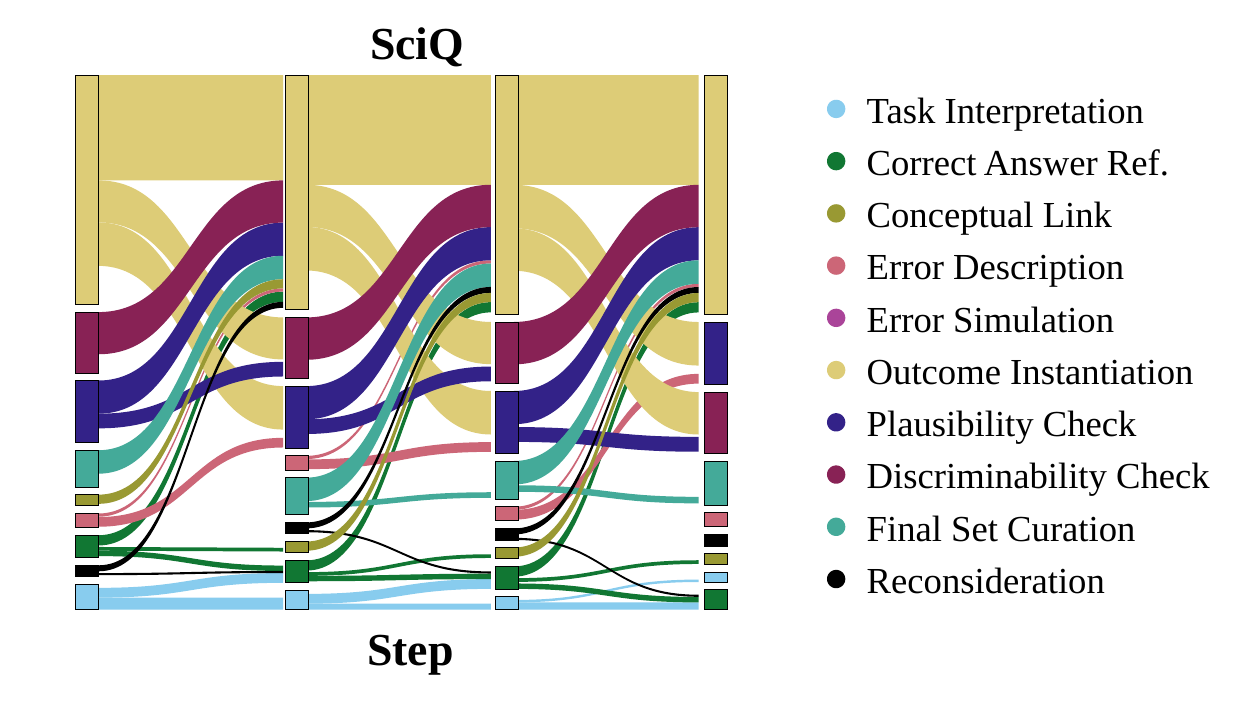}
    \end{minipage}
  \vspace{-7pt}
  \caption{
  Shows transition probabilities between strategies in \glm's reasoning traces. Sequences of strategies up to length 4 (left to right). Node height represents strategy share and link widths indicate the transition probabilities between successive strategies. Only dominant (>15\% outgoing mass) transitions are visualized for simplicity.\looseness=-1
  }
  \label{fig:glm-reasoning-transition-probabilities}
  \vspace{-10pt}
\end{figure*}

\cref{fig:deepseek-cot-transition-probabilities} shows common strategy transitions for \ds in the CoT setting, while~\cref{fig:glm-cot-transition-probabilities} and~\cref{fig:glm-reasoning-transition-probabilities} show the common transitions for \glm in The CoT and reasoning settings respectively.
Across models and settings, self-transitions for \texttt{INTER} and \texttt{CORR} are frequent, suggesting that these strategies often span multiple segments.

To compare two settings (i.e., models or prompting settings), we compute an agreement score over the major dominant outgoing transitions: for each source strategy and step, we identify the top three transitions exceeding 15\% outgoing probability mass and compute the set overlap between \ds and \glm, normalized by the smaller set size.
Averaging over source strategies and steps yields an agreement of 0.90 when varying only models (and averaging over datasets), and 0.91 when varying only prompting settings (and averaging over models), indicating broadly consistent dominant transition structures.
This agreement score still permits several minor differences between settings:
(i) the ordering of outgoing dominant edges may differ; for example, on Eedi, the most dominant outgoing edge of \glm in CoT from \texttt{ERR\_DESC} is to \texttt{ERR\_SIM}, whereas for \ds it is to \texttt{ERR\_DESC}. The former corresponds to iteratively listing an error and directly simulating it, whereas the latter corresponds to listing multiple errors at once.
(ii) even when the ordering of dominant edges is consistent, the ratios between transition masses can differ; for instance on Eedi, for \texttt{INST} in CoT, the ratio of transitions to \texttt{ERR\_DESC} vs self-transitions is much larger for \glm than for \ds.
(iii) the number of dominant edges may vary; for example on Eedi, \texttt{CORR} has 4 dominant outgoing edges in \ds's reasoning traces compared to just 2 in \glm's.

\section{Details on Discussion Experiments}
\label{app:discussion-experiments}
This section provides a detailed description of the experiments performed as part of the discussion (\cref{sec:discussion}).

\subsection{Solve-First and Error Injection}
\label{app:solve-first-error-inject}
To measure how frequently the LLM first solves a problem correctly and then injects an error into the solution, we annotate the LLMs' traces at scale. Following the approach used in our main analysis, we first define the following taxonomy:
\begin{itemize}
    \item \texttt{SOLVE\_FIRST}: Does the model systematically solve the problem to reach the correct final answer?
    \item \texttt{INJECT\_FROM\_SOLUTION}: Does the model refer to the correct solution or intermediate steps to generate plausible errors, misconceptions, or distractors?
\end{itemize}
We limit the analysis to problems where the LLM generated at least two steps in a step-by-step solution, ensuring that error injection is possible. Step-by-step solutions are generated using the prompt in \cref{tab:math_solver_prompt}.
Validation is performed by measuring the agreement with manual annotation on 50 traces. This yields a  Cohen's $\kappa$ of $0.85$ ($p < 0.001$).
To identify similarity-based strategies, we manually inspect reasoning traces filtered for semantically similar distractor values.

We find that in 92.5\% of \ds traces and 97.8\% of \glm traces, the model constructs a full step-by-step solution before proposing distractors.
In 71.2\% (\ds) and 76.3\% (\glm) of these cases, the model diverges at a specific step to inject an error, closely mirroring misconception-based design.

\paragraph{Automatic annotation.} \gptfivetwo~\citep{singh2026openaigpt5card} is used to automatically annotate both \texttt{SOLVE\_FIRST} and \texttt{INJECT\_FROM\_SOLUTION} for all traces. Explanations or relevant excerpts from the distractor-generation trace are also extracted to support the assigned labels (prompt in \cref{tab:solve_first_inject_prompt}).

\paragraph{Annotation quality.} A stratified subset of 50 samples are labeled manually. Agreement between manual and automatic annotations yields $\kappa = 0.85$ ($p < 0.001$).

\begin{promptbox}
\textbf{System Prompt} \par
\promptrule
{
\ttfamily
You are a helpful assistant.
\par}
\promptrule
\textbf{User Prompt} \par
\promptrule
{
\ttfamily
Solve the math problem. Output format (exact):\par
{[}STEP-1{]}...\par
...\par
{[}STEP-N{]}...\par
{[}FINAL ANSWER{]}...\par

Constraints: no fluff; only one reasoning/arithmetic step at a time; show key work. Final answer MUST equal:\par
"\promptvar{\{final\_answer\}}"\par

Problem:\par
"\promptvar{\{problem\_formulation\}}"
\par}
\end{promptbox}
\promptcaption{Prompt for solving a math problem \promptvar{\{problem\_formulation\}} step-by-step, in a manner that is consistent with the ground truth answer \promptvar{\{final\_answer\}}.}{tab:math_solver_prompt}

\begin{promptbox}
\textbf{System Prompt} \par
\promptrule
{
\ttfamily
You will be given a reasoning trace whose purpose is to generate *plausible incorrect distractor answers* for a math problem (i.e., simulate student mistakes). Your job is to label two independent properties: \par

**1. SOLVE\_FIRST:** Does the trace *attempt to solve the original problem correctly step-by-step* (compute the correct answer path), even if it later discusses mistakes/distractors? \par

- YES if it tries to lay out a solution procedure and/or computes intermediate results toward the true answer. \par
- NO if it jumps straight to distractors/misconceptions without first doing a solution attempt. \par

**2. INJECT\_FROM\_SOLUTION:** Does the trace *use the correct solution path as scaffolding* by referencing a specific step/intermediate quantity from the correct solution and then ``branching off'' by applying an incorrect operation/value there, propagating that error to a final distractor? \par

- YES only if you can point to a concrete ``branch point'' tied to the correct solution (e.g., ``Up to step X it's correct; then it miscomputes X or applies the wrong operation to X, yielding distractor Y''). \par
- NO if distractors are proposed without an explicit branch point from the correct solution (e.g., ``a possible distractor is 30 m/s'' with no tie to a specific intermediate step), OR if it only lists misconceptions abstractly without anchoring them to a step in the correct solution. \par

These labels are **independent**: A can be YES while B is NO (solves correctly first, then proposes distractors without branching). B is unlikely if A is NO, but still label based on the text. \par

Return output in EXACTLY this format (for regex parsing): \par

SOLVE\_FIRST\_DISCUSSION: <short quote or paraphrase from trace that is relevant for SOLVE\_FIRST; or ``N/A''> \par
SOLVE\_FIRST\_LABEL: YES|NO \par
INJECT\_FROM\_SOLUTION\_DISCUSSION: <short quote or paraphrase from trace that is relevant for INJECT\_FROM\_SOLUTION; or ``N/A''> \par
INJECT\_FROM\_SOLUTION\_LABEL: YES|NO \par

**Mini-examples (for guidance only):** \par

Example 1 (SOLVE\_FIRST=YES, INJECT\_FROM\_SOLUTION=YES): \par
Reasoning Trace: ``Let's solve: 30min=0.5h; v=60/0.5=120km/h; /3.6=33.3m/s. Student might incorrectly compute 60/0.5 as 30km/h, then convert...'' \par
Expected Output: \par
``SOLVE\_FIRST\_DISCUSSION: Correct problem solving present. \par
SOLVE\_FIRST\_LABEL: YES \par
INJECT\_FROM\_SOLUTION\_DISCUSSION: Branches off at the 60/0.5 step with a calculation mistake. \par
INJECT\_FROM\_SOLUTION\_LABEL: YES'' \par

Example 2 (SOLVE\_FIRST=NO, INJECT\_FROM\_SOLUTION=NO): \par
Reasoning Trace: ``A potential distractor is 30 m/s.'' \par
Expected Output: \par
``SOLVE\_FIRST\_DISCUSSION: No problem solving attempted. \par
SOLVE\_FIRST\_LABEL: NO \par
INJECT\_FROM\_SOLUTION\_DISCUSSION: N/A \par
INJECT\_FROM\_SOLUTION\_LABEL: NO'' \par

Example 3 (SOLVE\_FIRST=YES, INJECT\_FROM\_SOLUTION=NO): \par
Reasoning Trace: ``The correct solution is: 30min=0.5h; v=60/0.5=120km/h; /3.6=33.3m/s. One potential distractor is 45m/s.'' \par
Expected Output: \par
``SOLVE\_FIRST\_DISCUSSION: Question is first solved correctly. \par
SOLVE\_FIRST\_LABEL: YES \par
INJECT\_FROM\_SOLUTION\_DISCUSSION: No solution step is referenced and no related branching off takes place. \par
INJECT\_FROM\_SOLUTION\_LABEL: NO'' \par

Now label the following reasoning trace. \par

$[$BEGIN TRACE$]$ \par
\promptvar{\{trace\}} \par
$[$END TRACE$]$
\par}
\end{promptbox}
\promptcaption{Prompt for annotating solve-first and inject-from-solution behavior in LLMs' traces \promptvar{\{trace\}} when generating distractors.}{tab:solve_first_inject_prompt}

\subsection{Error Simulation}
\label{app:error-simulation}
\paragraph{Simulation prompt.}
Errors are simulated using the prompt in \cref{tab:student_sim_prompt}, with \ds in reasoning mode.

\begin{promptbox}
\textbf{System Prompt} \par
\promptrule
{
\ttfamily
You will be given a math question and specific student error. Please generate the incorrect answer that a student would give on the current question if they made the specified error.

At the end, give the student's final concise answer preceded with `Incorrect Student Answer:'
\par}
\promptrule
\textbf{User Prompt} \par
\promptrule
{
\ttfamily
Question: \promptvar{\{problem\_formulation\}}\par
Student Error: \promptvar{\{error\}}
\par}
\end{promptbox}
\promptcaption{Prompt for generating an incorrect student answer conditioned on a specific misconception \promptvar{\{error\}} on one math word problem \promptvar{\{problem\_formulation\}}.}{tab:student_sim_prompt}

\paragraph{Manual analysis for consistency.}
To evaluate consistency with any reasonable interpretation of the provided error, we first compare the generated answer against all available distractors for the same question that are sourced from the same error. For cases where the answer does not match any available distractor, we inspect the reasoning trace step-by-step to verify whether the error was correctly implemented. Albeit rarely the case, we consider examples consistent if the error is correctly implemented but is combined with a second error. Most instances are straightforward to evaluate; for example, errors such as ``rounds too much'' are trivial to classify.

\paragraph{Trace level analysis.} We analyze the reasoning traces for whether they contain the correct step by step solution to the problem, whether the misconception is discussed right in the beginning, after the correct solution, or injected midway at the relevant step, the dominant stance used by the LLM (student vs. expert), etc. We annotate these aspects for each trace using \gptfourone (see \cref{tab:error_sim_analysis_prompt} for the prompt used), and perform a manual verification of the annotations on a subset of 20 traces (Cohen's $\kappa$: 0.804 for \texttt{application\_pattern}, 0.894 for \texttt{correct\_path\_present}, 0.898 for \texttt{dominant\_stance}, and 1.0 for the rest of the tags, $p$ < 0.001 for all tags). We find that 68.3\% traces contained the correct solution to the underlying problem. 64.34\% traces first produced the correct solution, then introduced the misconception, whereas 30.3\% discuss applying the misconception from the start. \ds held a consistent expert stance in almost all (99.3\%) traces (of which, 66.9\% third person and 32.4\% first person). We also found that valid traces, i.e., resulting in the desired misconception-grounded distractor, were significantly shorter (based on a Mann-Whitney U test; $U$: 5728, $p$<0.05) in terms of average step count (3.377 steps) compared to invalid traces (3.972 steps).

\begin{promptbox}
\textbf{System Prompt} \par
\promptrule
{
\ttfamily
You are analyzing a chain-of-thought trace produced by an LLM that was asked to simulate a wrong answer a student would give to a math question, conditioned on a specific student misconception. Your job is to characterize *how* the trace handles the misconception — not to grade the math.
You will be given:
- question: the math question shown to the simulator\par
- correct\_answer: the ground-truth answer\par
- misconception: the high-level error the simulator was told to apply\par
- sim\_answer: the distractor the simulator produced\par
- sim\_reasoning: the simulator's full chain-of-thought\par
Read sim\_reasoning carefully and assign the following labels.\par
\textbf{correct\_path\_present}\par
  True iff the trace explicitly derives or states the correct answer at any point,independent of the misconception. Mentioning the correct method without computing it counts as False.\par
\textbf{application\_pattern}\par
  The dominant structural shape of the trace.\par
  - correct\_then\_corrupt\_full : correct answer derived in full, then misconception introduced and the erroneous path re-executed end-to-end\par
  - correct\_then\_corrupt\_skip : correct answer derived in full, then misconception introduced and the final wrong answer stated without re-executing the steps\par
  - corrupt\_partial : correct path begun, misconception introduced mid-derivation, execution continues from that point with the corrupted step\par
  - corrupt\_first : misconception applied from the start; correct path absent or only acknowledged in passing\par
  - other
\textbf{misconception\_evidence}\par
   1–3 short verbatim snippets from sim\_reasoning that justify the labels above in relation to where and how the misconception is induced in sim\_reasoning; each <= 25 words.\par
\textbf{dominant\_stance}\par
    The overall narrative perspective of the trace.\par
   - expert\_third\_person : the model speaks *about* a student in the third person ("the student would do X", "they might think Y"). Pedagogical / diagnostic framing.\par
   - first\_person\_student : the model speaks *as* the student, in the first person ("I think...", "I'll multiply...", "my answer is..."), inhabiting the role.\par
   - first\_person\_solver : the model uses first-person language but as itself solving / deciding ("I need to compute...", "I'll go with 1 as the answer"), not roleplaying the student. This is the AI narrating its own task execution.\par
   - impersonal : neither perspective is marked — the trace is mostly bare computation, passive voice, or pronoun-free statements ("4 × -3 = -12. Therefore the answer is -12.").\par
   - mixed : two or more stances are roughly comparable in prominence and neither dominates.\par
\textbf{stance\_shift}\par
   Does the stance change across the trace?\par
   - stable : one stance throughout\par
   - shifts\_to\_student : starts expert/solver/impersonal, moves into first-person student voice at some point\par
   - shifts\_from\_student : starts in student voice, moves out of it\par
   - oscillates : repeatedly switches between stances\par
   - n\_a : trace is too short or too uniform to assess\par
\textbf{student\_voice\_present}\par
   True iff the trace contains at least one clear instance of the model speaking *as* the student in the first person (not merely first-person solver language).\par
\textbf{expert\_voice\_present}\par
   True iff the trace contains at least one clear instance of expert/third-person framing about the student ("the student would...", "they might...", "a learner
   who believes X...").\par
\textbf{pedagogical\_framing}\par
   True iff the trace explicitly reasons *about* the misconception as a teaching / diagnostic object (e.g. naming it, classifying it, discussing what kind of
   student holds it, comparing it to other errors) rather than just applying it.\par
\textbf{stance\_evidence}\par
   2–4 short verbatim snippets from sim\_raw or sim\_reasoning that justify the labels above in relation to the stance used; each <= 25 words. Include at least one snippet for each stance you marked as present.\par
\par}
\promptrule
\textbf{User Prompt} \par
\promptrule
{
\ttfamily
question: \promptvar{\{problem\_formulation\}}\par
correct\_answer: \promptvar{\{correct\_answer\}}\par
misconception: \promptvar{\{student\_error\}}\par
sim\_answer: \promptvar{\{sim\_answer\}}\par
sim\_reasoning: \promptvar{\{sim\_reasoning\}}\par
\par}
\end{promptbox}
\promptcaption{Prompt for annotating error simulation reasoning traces.}{tab:error_sim_analysis_prompt}

\subsection{Enforcing a Literature-Informed Reasoning Structure}
\label{app:literature-informed}

\paragraph{Literature-informed prompt.}
To test whether models benefit from being explicitly instructed to follow the misconception-based procedure, we use the prompt in \cref{tab:ls_informed_distractor_prompt}, which enforces the full reasoning structure: solve the problem first, enumerate common error primitives, simulate each error, assess plausibility and discriminability, and select the final set. \Cref{tab:model-diversity-ls} reports proportional match under this literature-informed prompt (\emph{w/ LS}) against the neutral prompt (\emph{w/o LS}).

\begin{promptbox}
\textbf{System Prompt} \par
\promptrule
{
\ttfamily
You will be given a math question. Please generate \{n\} incorrect distractor answers for the question to be used as multiple-choice options in a multiple-choice exam.

**RULES** \par
Solve correctly first: \par
1. First, solve the problem correctly and treat the most likely correct solution as fixed. \par
2. Identify the key concepts involved in the correct solution. \par
3. Specify the exact conditions that an answer must violate to be a valid distractor. \par

Error modeling: \par
4. Enumerate at most 7 common error primitives relevant to this problem. Each error primitive must be either: \par
   (a) a buggy rule (an incorrect transformation or procedure), or \par
   (b) a buggy commitment (a false assumption or misclassification treated as true). \par
5. Each error primitive must be specific, stable, and capable of producing a concrete deterministic answer. \par

Error simulation: \par
6. For each error primitive, assume the student fully commits to that single error and reasons correctly in all other respects. \par
7. Derive the final incorrect answer that results from that single error and collect it as a distractor candidate. \par
    - If the question is non-numerical, output the final incorrect conclusion, classification or choice that follows from the error primitive. \par

Plausibility Assessment: \par
8. For each distractor candidate you assess its discriminative power: \par
    - check if the candidate can be derived deterministically \par
    - check if the candidate is truly incorrect under any reasonable interpretation \par
    - quantify how likely students are to stop here and select this candidate \par
    - make sure it is unambiguous and well-formed \par

Selection: \par
9. From the remaining candidates, select the \{n\} most distinct plausible distractors. \par

Output rules: \par
10. You should only output the final concise distractor values in the following template: \par

$[$Template$]$ \par
Distractor1: \par
... \par
Distractor\{n\}:
\par}
\promptrule
\textbf{User Prompt} \par
\promptrule
{
\ttfamily
Question: \promptvar{\{problem\_formulation\}}
\par}
\end{promptbox}
\promptcaption{Prompt for enforcing an ideal, learning-science-informed distractor generation procedure for a single math problem \promptvar{\{problem\_formulation\}}.}{tab:ls_informed_distractor_prompt}
\vspace{-10pt}

\paragraph{The two annotated models.}
For \ds\ and \glm, enforcing the structure yields no significant change ($p>0.05$, paired $t$-test; \cref{tab:model-diversity-ls}), consistent with our interpretation that they already employ a process closely resembling the misconception-based approach in their reasoning (\cref{sec:discussion}).

\paragraph{Generalization across models.}
\label{app:model-diversity}
The process-level analysis covers only \ds\ and \glm, as validating the automatic annotator is costly (discussed in limitations \cref{sec:limitations}). The \emph{performance} measurements, however, need no annotation, so we can report proportional match for a wider set of reasoning models and test whether the performance-level findings generalize. Beyond \ds\ and \glm\ we evaluate \gptoss, \gptosslarge, \gptfive, \glmflash, \gemma, and \geminis, all in reasoning mode, on the same $429$ solvable Eedi problems.

The clearest generalization is the strong absolute gap between math and science: on SciQ under the neutral prompt, proportional match stays in the $0.11$--$0.15$ band for every model we ran there (\gptoss $0.11$, \gptosslarge $0.12$, \glmflash $0.12$, \ds\ $0.14$, \glm\ $0.15$), far below the Eedi range, mirroring the Eedi/SciQ gap we analyze. Absolute Eedi performance itself varies across models, as expected for models of different families and sizes (see \cref{tab:model-diversity-ls}).

Enforcing the reasoning structure, by contrast, does not generalize uniformly. It produces no significant change for \ds, \glm, \gemma, and \geminis ($p>0.05$, paired $t$-test), consistent with our claim that these models already self-organize an effective reasoning structure (\cref{sec:discussion}). The models that improve are all from OpenAI: \gptoss, \gptosslarge, and \gptfive (all $p<0.01$; \cref{tab:model-diversity-ls}). These are also the lowest-scoring models under the neutral prompt, so model family and performance headroom are confounded and we do not attribute the gain to either. None of this bears on our claim, which concerns only the two models we annotate in depth: neither improves under enforced structure, consistent with the annotation finding that their traces already contain the misconception-based steps (\cref{sec:discussion}). \glmflash is the only model for which the neutral prompt scores significantly higher, and even there the effect is borderline. Under the structure-enforcing prompt it frequently falls into repetition loops, so we attribute its drop to a failure to follow the multi-step instruction rather than to a cost of the structure itself.

\begin{table*}[!htbp]
\centering
\renewcommand{\arraystretch}{1.15}
\small
\begin{tabular}{ll|cc}
\toprule
\textbf{Model} & \textbf{Size} & \textbf{w/o LS} & \textbf{w/ LS} \\
\midrule
\ds          & 685b & $0.52 \pm 0.03$ & $0.55 \pm 0.03$ \\
\glm         & 358b & $0.52 \pm 0.03$ & $0.52 \pm 0.03$ \\
\midrule
\gptoss      & 20b  & $0.30 \pm 0.03$ & $\mathbf{0.35 \pm 0.03}$ \\
\gptosslarge & 120b & $0.36 \pm 0.03$ & $\mathbf{0.46 \pm 0.03}$ \\
\gptfive     & --   & $0.47 \pm 0.03$ & $\mathbf{0.52 \pm 0.03}$ \\
\midrule
\glmflash    & 31b  & $\mathbf{0.40 \pm 0.03}$ & $0.37 \pm 0.03$ \\
\gemma       & 31b  & $0.57 \pm 0.03$ & $0.55 \pm 0.03$ \\
Gemini-2.5-flash & -- & $0.53 \pm 0.03$ & $0.54 \pm 0.03$ \\
Gemini-2.5-pro   & -- & $0.52 \pm 0.03$ & $0.53 \pm 0.03$ \\
\bottomrule
\end{tabular}
\caption{Proportional match on the $429$ solvable Eedi problems (reasoning mode) under the neutral prompt (\emph{w/o LS}) and the literature-informed prompt enforcing the full reasoning structure (\emph{w/ LS}). The two models we annotate in depth (top block) show no significant change from enforced structure; among the additional models, only the OpenAI models (\gptoss, \gptosslarge, and \gptfive) improve significantly. These improvers are also among the lowest-scoring models under the neutral prompt, so a model-family effect and a performance-headroom effect are confounded here and we do not attribute the gain to either. \textbf{Bold} marks the setting that is significantly better per model (paired $t$-test, $p<0.05$); non-bold pairs do not differ significantly.}
\label{tab:model-diversity-ls}
\end{table*}
\subsection{Oracle Ranker over Proposed Candidates}
\label{app:oracle-ranker}
The oracle ranker isolates how much of the selection loss (\cref{sec:discussion}) comes from the model discarding a valid candidate rather than never proposing it. We count a human-authored distractor as recovered if the model instantiates an equivalent value anywhere in its reasoning, not only in its final three.

\paragraph{Oracle proportional match.}
For each problem we build a candidate pool: the model's final selected distractors, together with every candidate it proposes elsewhere in its reasoning (recovered from the \texttt{INST} spans; details below). We call a human-authored distractor \emph{matched} if some pool member is judged equivalent to it, using the same \gptfourone\ judge as in \cref{app:measure-performance}. The oracle proportional match is the fraction of human-authored distractors matched this way, counting each at most once. The actual proportional match is the same fraction over the final selected distractors alone. The oracle is therefore always at least the actual, and their difference counts the distractors the model proposed but did not select.

\paragraph{Recovering proposed candidates from \texttt{INST}.}
To recover the proposed candidates we use the \texttt{INST} tags, which mark where the model instantiates a concrete candidate distractor. For every \texttt{INST} occurrence we reconstruct a short excerpt of the 220 characters preceding the tag, so that the excerpt terminates exactly at the proposed value. We then prompt \gptfourone\ (greedy decoding) to extract the concrete answer value at the end of the excerpt (\cref{tab:oracle-extraction-prompt}); the question and correct answer are supplied for context. We validated the extraction on $100$ held-out excerpts sampled uniformly across datasets, models and CoT vs reasoning setting. We manually judge each extracted value against the excerpt: $98\%$ were correct.

\paragraph{Population estimates via inverse-probability weighting.}
Because the annotated traces are stratified on performance (\cref{app:behaviour-performance}), a raw mean over them does not estimate the population proportional match. To report population-level figures we inverse-probability-weight each trace by the inverse of its within-bucket sampling rate, the population count of its bucket divided by the number of traces sampled from it. The two buckets are solvable traces with proportional match ${>}0.5$ and ${<}0.5$. For \ds\ this recovers the reported full-data proportional match: $0.51$ on Eedi, within the reported $0.52 \pm 0.03$, and $0.14$ on SciQ, matching the reported value exactly. \Cref{tab:oracle-match} reports the population-weighted oracle and actual proportional match for all eight dataset$\times$model$\times$prompting-setting combinations.
Note how the paired $t$-test needs no weighting because it is a within-trace comparison: each trace contributes its own oracle-minus-actual difference, and the test asks only whether that difference is reliably positive. Weighting matters solely for projecting the stratified sample onto a population average (the point estimates), not for a difference evaluated separately within each trace, where the oracle is by construction never below the actual.

\begin{promptbox}
\textbf{System Prompt} \par
\promptrule
{
\ttfamily
You are given an excerpt from an LLM's reasoning while it invents wrong answer options (distractors) for a multiple-choice question. The excerpt ends with a <INST> marker. Immediately to the LEFT of that marker is one candidate distractor value that the reasoning just named. Your job is purely positional: output the value token(s) directly before <INST>, not any other value appearing earlier in the excerpt. Clean it up: strip surrounding markdown/formatting, quotes, enumeration or field labels (e.g.\ `Distractor1:', `Idea 3:', `3.', `F ='), and any trailing bracketed or parenthetical comment. Use the question and correct answer only to judge where the value begins. Output ONLY that value, or exactly NONE if there is no concrete answer value immediately before the marker.
\par}
\promptrule
\textbf{User Prompt} \par
\promptrule
{
\ttfamily
<question>\promptvar{\{question\}}</question> \par
<correct\_answer>\promptvar{\{correct answer\}}</correct\_answer> \par
<excerpt>\promptvar{\{$\sim$220-char excerpt ending at the candidate\}}<INST></excerpt>
\par}
\end{promptbox}
\promptcaption{Prompt used to extract the concrete candidate distractor value marked by each \texttt{INST} span (\cref{app:oracle-ranker}). The \texttt{<INST>} marker is re-appended at the end of the excerpt so the extractor targets the value the tag annotates rather than an earlier-mentioned candidate.}{tab:oracle-extraction-prompt}

\begin{table*}[!htbp]
\centering
\renewcommand{\arraystretch}{1.15}
\small
\begin{tabular}{ll l|c c c}
\toprule
\textbf{Dataset} & \textbf{Model} & \textbf{Setting} & \textbf{Actual} & \textbf{Oracle} & \textbf{$p$} \\
\midrule
Eedi & \ds  & reasoning & $0.51$ & $0.67$ & $7.9{\times}10^{-12}$ \\
Eedi & \ds  & CoT       & $0.51$ & $0.57$ & $1.1{\times}10^{-4}$ \\
Eedi & \glm & reasoning & $0.53$ & $0.70$ & $1.5{\times}10^{-11}$ \\
Eedi & \glm & CoT       & $0.46$ & $0.52$ & $5.5{\times}10^{-5}$ \\
\midrule
SciQ & \ds  & reasoning & $0.14$ & $0.29$ & $9.8{\times}10^{-8}$ \\
SciQ & \ds  & CoT       & $0.14$ & $0.19$ & $1.0{\times}10^{-3}$ \\
SciQ & \glm & reasoning & $0.19$ & $0.29$ & $5.7{\times}10^{-7}$ \\
SciQ & \glm & CoT       & $0.16$ & $0.20$ & $2.2{\times}10^{-3}$ \\
\bottomrule
\end{tabular}
\caption{Oracle vs.\ actual proportional match for all eight dataset$\times$model$\times$prompting-setting combinations. The \emph{oracle} counts a human-authored distractor as recovered whenever the model instantiates an equivalent value anywhere in its reasoning; the \emph{actual} counts only the final selected set. Both cap each human-authored distractor at one match and are inverse-probability-weighted to the population (\cref{app:oracle-ranker}); $p$ is the per-trace paired $t$-test (unaffected by weighting), and $n$ is $120$ per Eedi cell and $77$--$88$ per SciQ cell. The oracle exceeds the actual in every cell.}
\label{tab:oracle-match}
\vspace{-10pt}
\end{table*}

\subsection{Valid Distractors Dropped by Self-Checks}
\label{app:self-check-drops}
The oracle analysis (\cref{app:oracle-ranker}) shows that models frequently propose a valid distractor and then discard it. Here we ask how much of that loss the model's own plausibility and discriminability checks (\texttt{PLAUS}, \texttt{DISCR}) account for: we quantify how often such a check argues against keeping a candidate that matches a human-authored one.

\paragraph{Linking checks to candidate and sentiment.}
To count these cases we must know which candidate each check assesses, so we link every \texttt{PLAUS}/\texttt{DISCR} check to an \texttt{INST} candidate.
Our annotations already mark both the check tag and the \texttt{INST} candidates, but not which candidate a given check refers to. Because the tag sits at the end of the assessment and the candidate value can lie before or after it,\footnote{For instance, the value may sit to the left of the tag (``\dots giving $3600$~\texttt{<INST>}, but this is too close to the correct total~\texttt{<DISCR>}'') or to its right (``this is a tempting slip~\texttt{<PLAUS>}, giving $3600$~\texttt{<INST>}'').} we take a window of ${\sim}300$ characters to the check's left and ${\sim}160$ to its right. Within this window we strip all other tags and replace the check with a \texttt{{<}{<}{<}CHECK{>}{>}{>}} marker, then prompt \gptfourone\ to
(i) extract the single \texttt{INST} candidate value the check assesses, and (ii) classify its \emph{sentiment toward keeping that candidate} as positive, negative, or neutral (full prompt in \cref{tab:self-check-prompt}). Some checks target the distractor set as a whole or make a generic remark; for these the extractor returns \texttt{NONE}, and we exclude them from the drop-rate analysis.
Sentiment is \emph{negative} when the check argues against using the candidate, for example when the model judges it indistinguishable from the correct answer, too obvious or too obscure to be tempting, or an invalid answer form; \emph{positive} when it argues for keeping it; and \emph{neutral} when the remark is descriptive or undecided. We validated the joint extractor and classifier against manual judgements on $100$ held-out checks ($50$ \texttt{PLAUS}, $50$ \texttt{DISCR}) sampled across datasets, models, and prompting settings, reaching ${\sim}93\%$ agreement.

\begin{promptbox}
\textbf{System Prompt} \par
\promptrule
{
\ttfamily
You are analyzing an excerpt from an LLM inventing wrong answer options (distractors) for a multiple-choice question. The marker {<}{<}{<}CHECK{>}{>}{>} is a \promptvar{\{check type\}} check the model applied. The check's evaluative clause ENDS at the marker, so read the text immediately to the LEFT of {<}{<}{<}CHECK{>}{>}{>} to understand what the check concludes. Base BOTH the value and the verdict on that left clause; text AFTER the marker is the next reasoning step and must not override it. \par

STEP 1 -- VALUE: identify the single candidate distractor value that this check is assessing. Use the left clause to decide WHICH candidate is meant; the candidate's VALUE may appear either just before (left of) or just after (right of) the marker -- extract it from whichever side it appears. Output it verbatim, cleaned of labels/markup/quotes. Numbers or expressions that are part of the QUESTION itself (given terms, the sequence in the prompt, the correct answer being noted as correct) are NOT the assessed distractor value. \par

OUTPUT NONE (do not pick any single value) when the check does not target exactly ONE candidate, i.e. any of: \par
* it evaluates the distractor SET or SEVERAL candidates together -- tell-tale words `these', `they', `all', `both', `the three': e.g. `these are plausible', `these are reasonable errors', `these are all incorrect', `these are distinct', `all close enough to the correct answer', `these look good'. This applies even when specific values appear earlier in the excerpt: if the clause ENDING AT the marker is a collective judgement -- including a list of several values followed by one verdict (`630, 600 and 700 are plausible', `Incorrect: 48, 190, 112. These are plausible mistakes') -- output NONE, do NOT pick one of them; \par
* it is a general/procedural/meta remark (`I need three distinct distractors', `let me finalize', `so any method that does not yield X is incorrect'); \par
* it verifies or restates the CORRECT answer (`X is correct', `the logic holds', `Tom is correct'). \par
Only output a value when the check clearly assesses exactly ONE specific candidate distractor. \par

ALWAYS give a SENTIMENT in STEP 2, even when VALUE is NONE: judge the check's stance toward the distractor(s) it discusses. A collective endorsement (`these are plausible', `these are good distractors', `plausible but clearly wrong') is POSITIVE; a collective rejection (`these are too obvious') is NEGATIVE. \par

\textbf{Sentiment rule, plausibility variant (\texttt{PLAUS}):} \par
STEP 2 -- SENTIMENT: what is this plausibility check's SENTIMENT toward keeping the distractor? \par
A good distractor must be TEMPTING -- a wrong answer a student might actually pick. \par
- POSITIVE if the check endorses it: plausible/tempting/believable/a strong distractor, OR names any real reason a student could be tempted (a shared property, a common confusion, `sounds like', `related to', `contains X') -- even if the same sentence also notes it is ultimately wrong. A distractor being INCORRECT is expected and is NOT negative. \par
- NEGATIVE if the check argues against using it: not tempting, too obvious, too far-fetched/obscure, `unlikely', `not plausible', `a student who computes would see it's wrong', `too easy to rule out', `less common'/`rare', OR the model compares candidates and leans AGAINST this one (favours another over it). \par
- NEUTRAL if purely descriptive or undecided (`but maybe...'). \par

\textbf{Sentiment rule, discriminability variant (\texttt{DISCR}):} \par
STEP 2 -- SENTIMENT: what is this discriminability check's SENTIMENT toward keeping the distractor? \par
A good distractor must be clearly WRONG yet still worth offering (not secretly correct, not so obvious no one picks it). \par
- POSITIVE if the check endorses it as a good distractor: clearly/unambiguously WRONG yet usable. Being wrong is GOOD for a distractor, so `plausible but clearly wrong upon careful checking', `not the first/main/primary X', `the reverse process', `not directly Y', or simply the wrong answer => POSITIVE. \par
- NEGATIVE if the check argues against using it: it might actually be CORRECT or is indistinguishable from / equal to the correct answer (`that's correct, so not a distractor'), too close/ambiguous, an invalid answer form, so obviously wrong no one would pick it, OR the model rejects/sets it aside (`not exactly', `doesn't fit', `less common', `let's avoid'). \par
- NEUTRAL if purely descriptive or undecided (`but maybe the question expects...'). \par

Output exactly two lines: \par
VALUE: <value or NONE> \par
SENTIMENT: <POSITIVE|NEGATIVE|NEUTRAL>
\par}
\promptrule
\textbf{User Prompt} \par
\promptrule
{
\ttfamily
<question>\promptvar{\{question\}}</question> \par
<correct\_answer>\promptvar{\{correct answer\}}</correct\_answer> \par
<excerpt>\promptvar{\{excerpt with {<}{<}{<}CHECK{>}{>}{>} marker\}}</excerpt>
\par}
\end{promptbox}
\promptcaption{Prompt used to (i) extract the single \texttt{INST} candidate value assessed by each \texttt{PLAUS}/\texttt{DISCR} self-check and (ii) classify its sentiment toward keeping that candidate (\cref{app:self-check-drops}). The \promptvar{\{check type\}} token and the STEP~2 sentiment rule take the plausibility variant for a \texttt{PLAUS} check and the discriminability variant for a \texttt{DISCR} check; the rest of the prompt is shared.}{tab:self-check-prompt}

\paragraph{Drop rates.}
Each check is now linked to a candidate and a sentiment. We label the candidate \emph{valid} if its value is judged equivalent to some human-authored distractor and \emph{invalid} otherwise.\footnote{Using the same \gptfourone\ judge as for proportional match (\cref{app:measure-performance}).} For each dataset$\times$model$\times$prompting-setting combination we report the fraction of \texttt{PLAUS}/\texttt{DISCR} checks on valid candidates that are negative (the \emph{valid drop rate}), alongside the same fraction over checks on invalid candidates (the \emph{invalid drop rate}). We report rates rather than counts because the model proposes many more invalid than valid candidates. Rates are inverse-probability-weighted to the population using the same per-bucket weights as in \cref{app:oracle-ranker} (each check inherits its trace's weight).
\Cref{tab:self-check-drops} shows the results. The valid drop rate is below the invalid drop rate in every setting: for \ds\ reasoning the invalid drop rate is $1.9\times$ the valid rate on Eedi ($53\%$ vs.\ $28\%$) and $1.5\times$ on SciQ ($23\%$ vs.\ $15\%$), so the checks discriminate, rejecting a given invalid candidate more often than a given valid one.

This discrimination is only relative, however: in absolute terms a non-trivial share of the checks on valid candidates are still negative (up to $28\%$ on Eedi reasoning). The pattern is consistent across both models: while \glm's valid and invalid drop rates are both lower than \ds's in every reasoning cell, the ordering (invalid $>$ valid in every cell, Eedi $>$ SciQ, reasoning $>$ CoT) is identical. The CoT rows carry too few checks on valid candidates to be reliable (valid $n\leq7$), so the cross-model comparison above rests on the reasoning rows.
Split by check type, the two are negative at comparable rates (on Eedi reasoning, \ds: $29\%$ of \texttt{PLAUS} and $27\%$ of \texttt{DISCR} checks on valid candidates).
A negative check does not always translate into a drop: the model may keep the candidate anyway, or discard it for unrelated reasons. Of the checks on valid candidates, the share that are negative \emph{and} whose candidate is absent from the final three is a lower bound on the loss these checks drive: $19\%$ on Eedi and $10\%$ on SciQ for \ds\ reasoning.

\begin{table*}[!htbp]
\centering
\renewcommand{\arraystretch}{1.15}
\small
\begin{tabular}{ll l|c c}
\toprule
\textbf{Dataset} & \textbf{Model} & \textbf{Setting} & \textbf{Valid dropped} & \textbf{Invalid dropped} \\
\midrule
Eedi & \ds  & reasoning & $28\%$ ($n{=}69$) & $53\%$ ($n{=}386$) \\
Eedi & \ds  & CoT       & $10\%$ ($n{=}4$)  & $64\%$ ($n{=}56$) \\
Eedi & \glm & reasoning & $17\%$ ($n{=}34$) & $34\%$ ($n{=}118$) \\
Eedi & \glm & CoT       & $10\%$ ($n{=}4$)  & $43\%$ ($n{=}44$) \\
\midrule
SciQ & \ds  & reasoning & $15\%$ ($n{=}30$) & $23\%$ ($n{=}238$) \\
SciQ & \ds  & CoT       & $8\%$ ($n{=}7$)   & $9\%$ ($n{=}30$) \\
SciQ & \glm & reasoning & $6\%$ ($n{=}20$)  & $17\%$ ($n{=}228$) \\
SciQ & \glm & CoT       & $0\%$ ($n{=}2$)   & $10\%$ ($n{=}48$) \\
\bottomrule
\end{tabular}
\caption{Fraction of \texttt{PLAUS}/\texttt{DISCR} self-checks that express \emph{negative} sentiment (arguing against keeping the candidate), split by whether the check's target candidate is \emph{valid} (equivalent to a human-authored distractor) or \emph{invalid}. Inverse-probability-weighted to the population (\cref{app:oracle-ranker}); $n$ is the raw count of checks per group. Checks on valid candidates are negative far less often than checks on invalid ones (the checks discriminate) but still at a non-trivial rate. CoT valid counts are small and are shown for completeness only.}
\label{tab:self-check-drops}
\end{table*}

\subsection{Strategy Usage and Performance}
\label{app:behaviour-performance}
To test whether the strategies in our taxonomy are correlated with output quality, we ask, in two stages, whether and how strategy usage differs between traces with prop match > 0.5 (high performance) and prop match < 0.5 (low performance).

\paragraph{Experimental setup.}
Three choices apply to both stages: which traces we analyse, how we summarize each one, and how we test for a difference. Our annotated traces are a stratified sample: per run, up to 60 traces with high performance and 60 with low performance, drawn from all solvable traces. The high-performance stratum reaches the full 60 per run on Eedi but is capped by availability on SciQ ($37$--$44$ per run; \cref{app:method-annotation}). Because we chose how many traces of each kind to annotate, high- and low-scoring traces appear here in near-equal numbers, which is not their ratio in the full data. Any measure that depends on that ratio is therefore distorted, including a correlation between strategy usage and the continuous proportional-match score. An average taken inside the high-scoring group, or inside the low-scoring group, does not depend on it. Both stages therefore compare the two groups directly rather than relating strategy usage to the continuous score, and neither needs reweighting.

We summarize each trace by its usage of the ten strategies in two ways: the absolute \emph{count} of each strategy, and its \emph{share}, the count as a percentage of the number of tags in the trace; a difference between two shares is therefore in percentage points (pp). A count grows as a trace gets longer; a share does not. Every $p$-value below comes from a permutation test with $200{,}000$ resamples, which needs no distributional assumptions: it reshuffles the high/low-scoring labels and asks how often chance alone reproduces a difference this large. \ds and \glm may use a given strategy at different rates overall, so we reshuffle labels only within a model; a difference between the two models can then never surface as a difference between high- and low-scoring traces. Both stages also control for problem difficulty, which we proxy by solution step count: the number of steps in a step-by-step reference solution generated for each Eedi problem, with the correct answer to the MCQ given as the required endpoint.

\paragraph{Stage 1: is the strategy profile different at all?}
Before comparing individual strategies, we test per dataset and prompting setting whether the ten-strategy usage profile differs between high- and low-scoring traces at all, using PERMANOVA \citep{anderson2001permanova,mcardle2001} on the Bray--Curtis dissimilarity of the count profiles. Each trace is one vector of ten strategy counts, Bray--Curtis measures how far apart two such vectors are, and PERMANOVA asks whether dissimilarity between the high- and low-scoring groups exceeds dissimilarity within them, with significance obtained by reshuffling the high/low-scoring labels. It is an omnibus test: it evaluates the whole profile at once rather than one strategy at a time, so it pays no multiplicity penalty, and gating on it means Stage 2 examines individual strategies only where there is evidence that something differs. On Eedi the profiles differ in both settings (reasoning $p=0.008$, CoT $p=0.032$; \cref{tab:behaviour-omnibus}).

Problem difficulty is a confounder here: if low-scoring traces come from harder problems, and difficulty shapes which strategies are used, the profile difference would be a property of the problems rather than of the traces. We therefore restrict the permutation to within terciles of solution step count, exchanging labels only between traces from problems of comparable difficulty. Under that restriction the reasoning $p$ is unchanged at $0.008$ and the CoT $p$ moves from $0.032$ to $0.034$, so the difference is not a difficulty artifact.

In contrast to Eedi, on SciQ no overall difference is detectable (reasoning $p=0.16$, CoT $p=0.06$); we therefore do not interpret per-strategy effects there.

\paragraph{Stage 2: which strategies drive the Eedi difference?}
Both Eedi settings pass Stage 1, so we ask which strategies drive the difference there. For each strategy we want the high$-$low difference in usage. As discussed in Stage 1, two factors could confound it: we pool two models whose baseline usage differs, and low-scoring traces may come from harder problems. A permutation test on its own gives a $p$-value but no difference adjusted for either. We therefore use an analysis of covariance, the standard method for comparing two groups while holding a covariate fixed. It is a linear model predicting a strategy's usage from three inputs: which model produced the trace, the problem's step count, and whether the trace is high- or low-scoring. Fitting means choosing one weight per input so that the predictions come as close as possible to the usage we observe. The weight on the high/low-scoring input is $\Delta$: the high$-$low difference that remains once model identity and step count are held fixed. We report it twice, without step count (\emph{naive}) and with it (\emph{step-adj.}). Because each model contributes 60 high-scoring and 60 low-scoring traces, the naive $\Delta$ reduces exactly to $\text{mean(high)}-\text{mean(low)}$. The linear model supplies the adjusted estimate only: every $p$-value below comes from a permutation test, so no distributional assumption is needed.

The $p$-value requires the same adjustment. Shuffling the high/low-scoring labels on their own would also break the association between usage and step count, and would therefore test a null hypothesis in which difficulty is randomized as well. A Freedman--Lane permutation \citep{freedman1983nonstochastic} avoids this: we refit the model without the high/low-scoring input, take the residuals (the usage left unexplained by model identity and step count), shuffle those, and refit. Only the high/low-scoring assignment is randomized, and the step-count structure is left in place. One further correction is needed because we test ten strategies per column: with ten independent tests the chance that at least one passes $p<0.05$ by chance is about 40\%, so we tighten the threshold with Benjamini--Hochberg \citep{benjamini1995controlling} and report $q$-values instead: at $q<0.05$, at most 5\% of the differences we mark significant are expected to be false. All $\Delta$ values quoted below are step-adjusted.

One strategy stands out (\cref{tab:behaviour-posthoc-eedi-reasoning,tab:behaviour-posthoc-eedi-cot}): error enumeration (\texttt{ERR\_DESC}) is more frequent in low-scoring traces under both prompting settings, in count and in share. Under reasoning the count difference is significant ($-15$ occurrences, $q=0.005$). The share difference is borderline significant in both settings ($-2.9$pp, $q=0.048$ under reasoning; $-4.2$pp, $q=0.044$ under CoT). Under CoT the count difference ($-29$ occurrences) is significant only before correction. Both models show the same sign for count and share, and step adjustment leaves the pattern unchanged, so problem difficulty does not explain the error-enumeration gap. Low-scoring traces also carry more tags overall, so a higher error-enumeration count alone could mean nothing more than more of everything; the share differences are what rule this out.

Two further share differences pass the correction: correct-answer reference (\texttt{CORR}, $+2.8$pp under CoT, $q=0.044$) and task interpretation (\texttt{INTER}, $+2.6$pp under reasoning, $q=0.024$) take a larger share of high-scoring traces without a matching rise in count, and the \texttt{CORR} count is significantly lower under reasoning. High-scoring traces carry roughly 50 fewer tags than low-scoring ones in both settings ($79$ vs $128$ per trace under reasoning, $23$ vs $73$ under CoT), and most of that gap is error enumeration and error simulation (\texttt{ERR\_SIM}), plus outcome instantiation (\texttt{INST}) under reasoning. A count that is unchanged in a trace with fewer tags is a larger share, which is all the \texttt{CORR} and \texttt{INTER} gains are. The reverse pattern appears for \texttt{ERR\_SIM} and \texttt{INST}: significant count gaps under reasoning that shrink or vanish as shares, so length accounts for those too. Read together, low-scoring traces are not drawn from harder problems, nor do they differ in which strategies appear; they carry more tags, and a larger part of those tags goes to enumerating student errors. Inflated error enumeration is the single behavioral signature that separates low- from high-scoring traces (\cref{sec:discussion}).

\paragraph{Remaining confounders.}
Both stages control for problem difficulty via solution step count and for trace length by reporting shares alongside counts. Other properties of a problem go uncontrolled and could contribute to the contrast.

One is the size of a problem's answer space. Some items admit very few candidate distractors. In the 37 problems of our annotated Eedi sample where two characters propose competing methods, the four options are always \emph{Only Tom}, \emph{Only Katie}, \emph{Both} and \emph{Neither}, so the three human-authored distractors are fixed in advance, a model recovers them easily, and proportional match saturates. Elsewhere the plausible answers are far more numerous. For $43.2 \div 10$ the human-authored distractors are $0.432$, $33.2$ and $43.02$, each encoding a different misconception, and a model can reach further errors that are equally plausible but match none of the three.

Step count does not capture this. Those same 37 problems have the smallest answer space in the sample yet the longest reference solutions, averaging $4.95$ steps against $3.44$ for the other 290; the one asking which of two proposed next lines of $\frac{x}{3}-5=12$ is correct takes eight steps to verify, while $43.2 \div 10$ takes two. Adjusting for step count therefore does not also adjust for answer-space size. If high- and low-scoring traces differ systematically in that size, and if it also shapes which strategies are used, part of the contrast would reflect problem type rather than reasoning quality.

Two further factors are uncontrolled: the topic of a problem within a dataset, such as algebra against geometry, and whether its human-authored distractors encode recognizable misconceptions, which error simulation can recover, or are arbitrary alternatives. We therefore read Stage 2 as a description of how high- and low-scoring traces differ rather than as an isolation of a causal effect of strategy choice on performance.

\begin{table}[!htbp]
\centering
\renewcommand{\arraystretch}{1.15}
\small
\begin{tabular}{ll|r r}
\toprule
\textbf{Dataset} & \textbf{Setting} & \textbf{pseudo-$F$} & \textbf{$p$} \\
\midrule
Eedi & reasoning & 3.95 & 0.008$^{*}$ (0.008) \\
Eedi & CoT & 2.60 & 0.032$^{*}$ (0.034) \\
SciQ & reasoning & 1.45 & 0.162 \\
SciQ & CoT & 2.10 & 0.058 \\
\bottomrule
\end{tabular}
\caption{Stage 1 omnibus: PERMANOVA (pseudo-$F$ on the Bray--Curtis dissimilarity of the ten-strategy count profile) for a difference between high-scoring (proportional match ${>}0.5$) and low-scoring (${<}0.5$) traces, pooling both models with restricted permutation within model. For Eedi the parenthesised $p$ is the difficulty-robust variant (permutation restricted within model$\times$step-count tercile). $^{*}p<0.05$.}
\label{tab:behaviour-omnibus}
\end{table}

\begin{table*}[!htbp]
\centering
\renewcommand{\arraystretch}{1.15}
\small
\begin{tabular}{l|r@{.}l r@{.}l|r@{.}l r@{.}l}
\toprule
& \multicolumn{4}{c}{\textbf{count}} & \multicolumn{4}{c}{\textbf{share (pp)}} \\
\cmidrule(lr){2-5}\cmidrule(lr){6-9}
\textbf{Strategy} & \multicolumn{2}{c}{naive} & \multicolumn{2}{c}{step-adj.} & \multicolumn{2}{c}{naive} & \multicolumn{2}{c}{step-adj.} \\
\midrule
Task Interpretation & $+0$ & $45$ & $+0$ & $48$ & $+2$ & $52^{*}$ & $+2$ & $57^{*}$ \\
Correct Answer Ref. & $-4$ & $24^{*}$ & $-4$ & $25^{*}$ & $+0$ & $01$ & $+0$ & $02$ \\
Conceptual Link & $-0$ & $07$ & $-0$ & $07$ & $-0$ & $12$ & $-0$ & $12$ \\
Error Description & $-15$ & $35^{**}$ & $-15$ & $42^{**}$ & $-2$ & $84^{\dagger}$ & $-2$ & $88^{*}$ \\
Error Simulation & $-11$ & $87^{*}$ & $-11$ & $94^{*}$ & $-1$ & $56$ & $-1$ & $57$ \\
Outcome Instantiation & $-15$ & $36^{*}$ & $-15$ & $45^{*}$ & $+0$ & $37$ & $+0$ & $36$ \\
Plausibility Check & $-1$ & $57$ & $-1$ & $57$ & $+0$ & $26$ & $+0$ & $27$ \\
Discriminability Check & $-0$ & $18$ & $-0$ & $18$ & $+0$ & $39$ & $+0$ & $39$ \\
Final Set Curation & $-0$ & $13$ & $-0$ & $13$ & $+0$ & $57$ & $+0$ & $57$ \\
Reconsideration & $-0$ & $62$ & $-0$ & $62$ & $+0$ & $41$ & $+0$ & $40$ \\
\bottomrule
\end{tabular}
\caption{Stage 2: high$-$low $\Delta$ in per-trace strategy usage for Eedi / reasoning traces (positive $=$ high-scoring traces use it more), pooling both models. \emph{count}: absolute occurrences; \emph{share}: percentage of the trace's tags. \emph{step-adj.}: adjusts for solution step count (Freedman--Lane). Permutation test, Benjamini--Hochberg across the ten strategies within each column ($^{*}q<0.05$, $^{**}q<0.01$; $^{\dagger}$ nominal $p<0.05$). The step-adjusted Error Description share is borderline ($q=0.048$).}
\label{tab:behaviour-posthoc-eedi-reasoning}
\end{table*}

\begin{table*}[!htbp]
\centering
\renewcommand{\arraystretch}{1.15}
\small
\begin{tabular}{l|r@{.}l r@{.}l|r@{.}l r@{.}l}
\toprule
& \multicolumn{4}{c}{\textbf{count}} & \multicolumn{4}{c}{\textbf{share (pp)}} \\
\cmidrule(lr){2-5}\cmidrule(lr){6-9}
\textbf{Strategy} & \multicolumn{2}{c}{naive} & \multicolumn{2}{c}{step-adj.} & \multicolumn{2}{c}{naive} & \multicolumn{2}{c}{step-adj.} \\
\midrule
Task Interpretation & $+0$ & $06$ & $+0$ & $04$ & $+1$ & $12$ & $+0$ & $95$ \\
Correct Answer Ref. & $-2$ & $58$ & $-2$ & $93$ & $+3$ & $49^{**}$ & $+2$ & $75^{*}$ \\
Conceptual Link & $+0$ & $01$ & $+0$ & $01$ & $+0$ & $05$ & $+0$ & $06$ \\
Error Description & $-27$ & $43^{\dagger}$ & $-28$ & $83^{\dagger}$ & $-4$ & $52^{*}$ & $-4$ & $15^{*}$ \\
Error Simulation & $-16$ & $73$ & $-17$ & $65$ & $-2$ & $25$ & $-1$ & $71$ \\
Outcome Instantiation & $-2$ & $28$ & $-2$ & $16$ & $+2$ & $66$ & $+2$ & $53$ \\
Plausibility Check & $-0$ & $44$ & $-0$ & $42$ & $-0$ & $32$ & $-0$ & $37$ \\
Discriminability Check & $-0$ & $33$ & $-0$ & $31$ & $+0$ & $29$ & $+0$ & $35$ \\
Final Set Curation & $-0$ & $17$ & $-0$ & $13$ & $-0$ & $33$ & $-0$ & $23$ \\
Reconsideration & $-0$ & $27$ & $-0$ & $24$ & $-0$ & $18$ & $-0$ & $17$ \\
\bottomrule
\end{tabular}
\caption{Stage 2: high$-$low $\Delta$ in per-trace strategy usage for Eedi / CoT traces (positive $=$ high-scoring traces use it more), pooling both models. \emph{count}: absolute occurrences; \emph{share}: percentage of the trace's tags. \emph{step-adj.}: adjusts for solution step count (Freedman--Lane). Permutation test, Benjamini--Hochberg across the ten strategies within each column ($^{*}q<0.05$, $^{**}q<0.01$; $^{\dagger}$ nominal $p<0.05$). The step-adjusted Error Description and Correct Answer Ref.\ shares are borderline ($q=0.044$ each).}
\label{tab:behaviour-posthoc-eedi-cot}
\end{table*}

\begin{figure*}[!htbp]
\centering
\includegraphics[width=\linewidth]{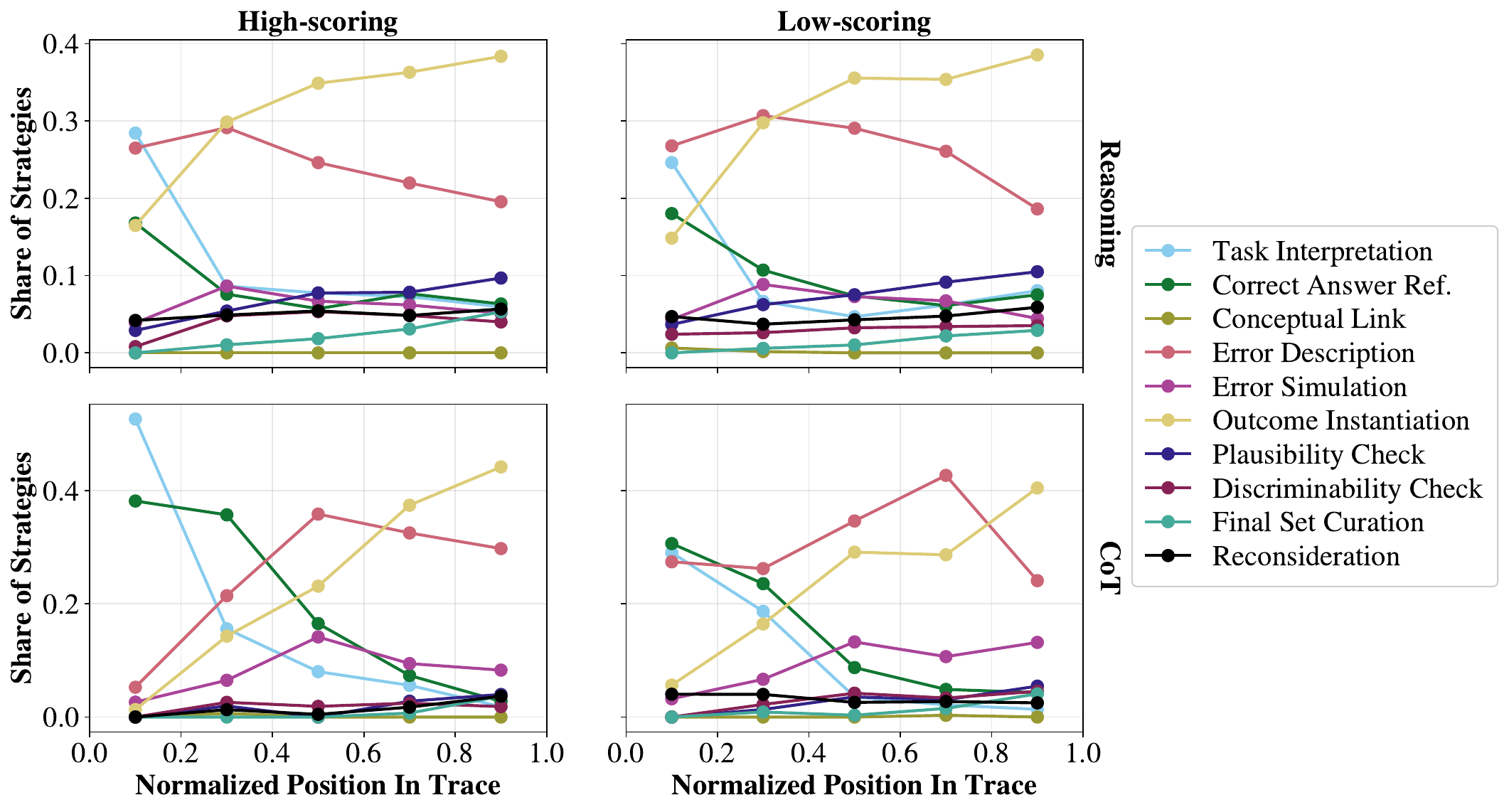}
\caption{Strategy usage over normalized trace position in \ds traces on Eedi, split by output quality (columns: high-scoring vs low-scoring, proportional match ${>}0.5$ vs ${<}0.5$) and prompting setting (rows). Within each fifth of the trace the strategy shares sum to one, as in \cref{fig:components-over-time}.}
\label{fig:perf-split-deepseek}
\end{figure*}

\begin{figure*}[!htbp]
\centering
\includegraphics[width=\linewidth]{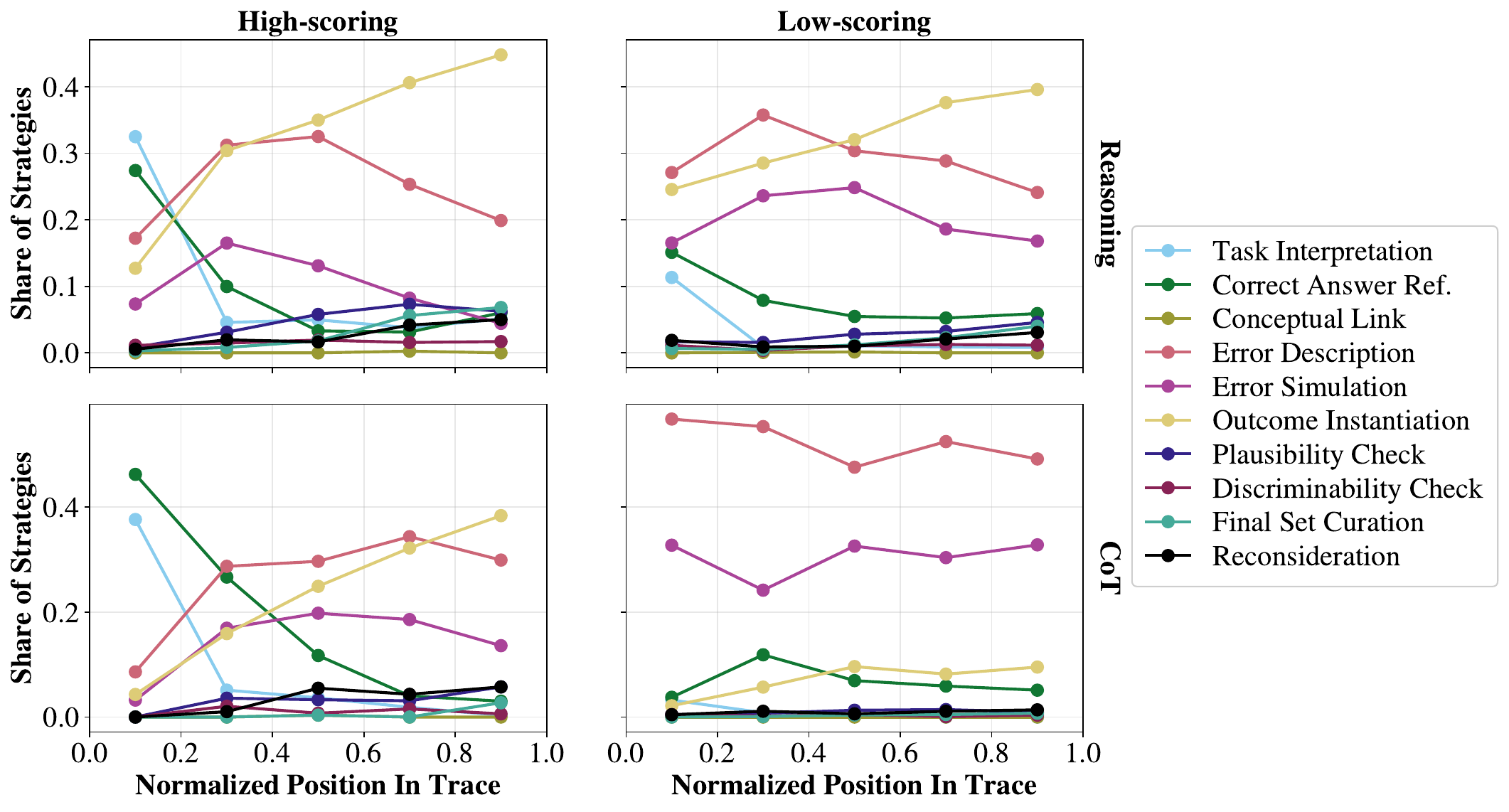}
\caption{Strategy usage over normalized trace position in \glm traces on Eedi, split by output quality and prompting setting, as in \cref{fig:perf-split-deepseek}.}
\label{fig:perf-split-glm}
\end{figure*}

\paragraph{Where in the trace high- and low-scoring traces diverge.}
Stage 2 says which strategies differ between high- and low-scoring traces, but not where in a trace the difference falls. \cref{fig:perf-split-deepseek,fig:perf-split-glm} show that, splitting the view of \cref{fig:components-over-time} by output quality for each model and prompting setting on Eedi. Each panel cuts a trace into five equal spans and gives the share of each strategy among the tags falling in one span, so it shows where a strategy is used rather than how much of it there is.

Under CoT the two groups differ most in the opening fifth. In high-scoring traces that span is almost all task interpretation (\texttt{INTER}) and correct-answer reference (\texttt{CORR}). In low-scoring traces error enumeration (\texttt{ERR\_DESC}) is already prominent there, where it is near-absent in high-scoring ones. \glm splits the same way on error simulation (\texttt{ERR\_SIM}); \ds does not. Under reasoning the shift appears in \glm but not in \ds, whose two panels are near-identical across all ten strategies. The CoT panels for high-scoring traces rest on few tags, since those traces are short: fewer than $100$ tags fall in the opening fifth for either model.

\section{Additional Experiments}

\subsection{Coverage of Strategies.}
Beyond the characterization of LLMs' distractor generation process in \cref{sec:experiments-process-analysis}, it is also informative to examine not only how frequently each strategy occurs on average within a trace, but also how many traces contain at least one instance of that strategy.

\begin{table*}[!htbp]
\centering
\renewcommand{\arraystretch}{1.15}
\small
\begin{tabular}{ll|
                r@{.}l r@{.}l
                r@{.}l r@{.}l}
\toprule
& &
\multicolumn{4}{c}{\textbf{CoT}}
& \multicolumn{4}{c}{\textbf{Reasoning}} \\
\cmidrule(lr){3-6}
\cmidrule(lr){7-10}

\textbf{Model} & \textbf{Strategy} &
\multicolumn{2}{c}{\textbf{Eedi}}
& \multicolumn{2}{c}{\textbf{SciQ}}
& \multicolumn{2}{c}{\textbf{Eedi}}
& \multicolumn{2}{c}{\textbf{SciQ}} \\
\midrule
\multirow{10}{*}{\textbf{\ds}}

& Task Interpretation
& 93 & 3
& 99 & 0
& 100 & 0
& 100 & 0 \\

& Correct Answer Ref.
& 98 & 9
& 100 & 0
& 100 & 0
& 100 & 0 \\

& Conceptual Link
& 1 & 1
& 94 & 8
& 1 & 7
& 99 & 0 \\

& Error Description
& 96 & 7
& 27 & 8
& 96 & 7
& 31 & 0 \\

& Error Simulation
& 43 & 9
& 0 & 0
& 55 & 6
& 0 & 0 \\

& Outcome Instantiation
& 97 & 8
& 95 & 9
& 100 & 0
& 100 & 0 \\

& Plausibility Check
& 65 & 0
& 84 & 5
& 96 & 7
& 91 & 0 \\

& Discriminability Check
& 20 & 0
& 75 & 3
& 54 & 4
& 93 & 0 \\

& Final Set Curation
& 57 & 2
& 85 & 6
& 93 & 9
& 100 & 0 \\

& Reconsideration
& 31 & 1
& 11 & 3
& 97 & 2
& 63 & 0 \\

\midrule

\multirow{10}{*}{\textbf{\glm}}

& Task Interpretation
& 65 & 0
& 100 & 0
& 98 & 3
& 100 & 0 \\

& Correct Answer Ref.
& 100 & 0
& 100 & 0
& 100 & 0
& 100 & 0 \\

& Conceptual Link
& 0 & 6
& 42 & 3
& 1 & 1
& 72 & 1 \\

& Error Description
& 98 & 3
& 57 & 7
& 100 & 0
& 81 & 7 \\

& Error Simulation
& 72 & 8
& 0 & 0
& 67 & 8
& 0 & 0 \\

& Outcome Instantiation
& 98 & 9
& 98 & 1
& 100 & 0
& 100 & 0 \\

& Plausibility Check
& 42 & 2
& 99 & 0
& 84 & 4
& 99 & 0 \\

& Discriminability Check
& 9 & 4
& 88 & 5
& 32 & 2
& 97 & 1 \\

& Final Set Curation
& 29 & 4
& 76 & 9
& 90 & 6
& 100 & 0 \\

& Reconsideration
& 31 & 1
& 5 & 8
& 71 & 1
& 60 & 6 \\

\bottomrule
\end{tabular}
\caption{Percentage of traces containing at least one occurrence of each taxonomy strategy (\cref{tab:taxonomy-definition}) for each model on both datasets.}
\label{tab:label_presence_comparison}
\end{table*}

\paragraph{Results.} In \cref{tab:label_presence_comparison}, we observe that across both \glm and \ds, reasoning traces show higher coverage of taxonomy strategies than CoT traces, especially for evaluative and iterative behaviors such as \texttt{PLAUS}, \texttt{CURATE}, and \texttt{RECON}. Core generative steps (e.g., \texttt{CORR} and \texttt{INST}) are nearly universal across models and prompting styles.
On Eedi, \glm in CoT shows less frequent usage of \texttt{LINK} but relatively higher usage of \texttt{ERR\_SIM} than \ds, indicating minor differences between models.

\end{document}